\PassOptionsToPackage{sort, compress}{natbib}

\documentclass[11pt, a4paper, colorlinks=true,allcolors=blue,hidelinks]{include/gdm_format}
\usepackage[numbers, sort&compress]{natbib}
\usepackage{amsmath,amsfonts,bm}

\def\eqref#1{equation~\ref{#1}}
\def\1{\bm{1}}

\DeclareMathAlphabet{\mathsfit}{\encodingdefault}{\sfdefault}{m}{sl}
\SetMathAlphabet{\mathsfit}{bold}{\encodingdefault}{\sfdefault}{bx}{n}

\usepackage{url}
\usepackage{booktabs} % For better table formatting
\usepackage{colortbl} % For coloring table cells
\usepackage{xcolor} % For defining colors
\usepackage{array} % For vertical centering
\usepackage{multirow} % For multirow cells
\usepackage{amssymb}
\usepackage{graphicx} % For including images
\usepackage{float}    % For [H] option to force the placement of the figure
\usepackage[most]{tcolorbox}
\usepackage{listings}
\usepackage{fontawesome5}
\usepackage{pgffor}
\usepackage{pgfplots}

\usepackage{amsmath}
\usepackage{amssymb} % For additional symbols
\usepackage{xspace}
\usepackage{multirow}
\usepackage{booktabs}
\usepackage{color}
\usepackage{colortbl}
\usepackage{cleveref}
\usepackage{graphicx}
\usepackage{algorithm}
\usepackage{algorithmicx}
\usepackage{algpseudocode}
\usepackage{subcaption}
\usepackage{wrapfig}
\usepackage{sidecap}
\usepackage{soul}
\usepackage{enumitem}
\sidecaptionvpos{figure}{t}
\usepackage{multicol}
\usepackage{pifont}
\usepackage[normalem]{ulem}
\usepackage{titletoc}
\usepackage{CJKutf8}
\usepackage{graphicx}
\usepackage{caption}
\usepackage{ulem}
\usepackage{nameref}
\newcolumntype{L}[1]{>{\raggedright\let\newline\\\arraybackslash\hspace{0pt}}m{#1}}
\newcolumntype{C}[1]{>{\centering\let\newline\\\arraybackslash\hspace{0pt}}m{#1}}
\newcolumntype{R}[1]{>{\raggedleft\let\newline\\\arraybackslash\hspace{0pt}}m{#1}}
\usepackage{tabularray}
\usepackage[export]{adjustbox}

 \usepackage{graphicx}
\usepackage{colortbl}
\usepackage{color}
\usepackage{footmisc}
\usepackage{enumitem}
\setlist{nolistsep}
\usepackage{xcolor}
\usepackage{diagbox}
\usepackage{makecell}
\usepackage{multirow}
\usepackage{booktabs}
\usepackage{geometry}
\usepackage{listings}
\usepackage{makecell}
\usepackage{multicol}
\usepackage{hyperref}
\usepackage{booktabs}
\usepackage{array}
\usepackage{marvosym} % For \Letter symbol
\usepackage{float}
\usepackage{pdfpages}
\usepackage{appendix}
\usepackage{balance}
\usepackage{xcolor}   
\usepackage{tcolorbox}

\usepackage{hyperref}
\usepackage{xurl}  % Allows URL breaking at any character
\usepackage{booktabs}
\usepackage{longtable}
\usepackage{supertabular}
\def\scititle{
Generalized Recognition of Basic Surgical Actions Enables Skill Assessment and Vision-Language-Model-based Surgical Planning}
\title{\bfseries \boldmath \scititle}
\usepackage{tikz}
\usetikzlibrary{fit}
\usetikzlibrary{positioning} % for [below = 2cm of node name]
\usetikzlibrary{calc}
\usepackage{pgfplots}
\pgfplotsset{compat=1.16}
\usepgfplotslibrary{fillbetween}
\usetikzlibrary{patterns}
\usepackage{siunitx}
\usepackage{pgf-pie}
\definecolor{OIblack}{RGB}{0, 0, 0}
\definecolor{OIgreen}{RGB}{0, 158, 115}
\definecolor{OIblue}{RGB}{0, 114, 178}
\definecolor{OIlightblue}{RGB}{86, 180, 233}
\definecolor{OIyellow}{RGB}{240, 228, 66}
\definecolor{OIorange}{RGB}{230, 159, 0}
\definecolor{OIred}{RGB}{213, 94, 0}
\definecolor{OIpink}{RGB}{204, 121, 167}
\usepackage{xspace}

\usepackage[acronym]{glossaries}
\glsdisablehyper
\newacronym{ras}{RAS}{robot-assisted surgery}
\newacronym{loa}{LoA}{level of autonomy}
\newacronym{srt}{SRT}{surgical robot transformer}
\newacronym{vlm}{VLM}{Vision-Language Model}
\newacronym{hl}{HL}{high-level}
\newacronym{ll}{LL}{low-level}
\newacronym{ce}{CE}{cross-entropy}
\newacronym{ood}{OOD}{Out-of-Distribution}
\newacronym{gui}{GUI}{graphical user interface}
\newacronym{mlp}{MLP}{multi-layer perceptron}
\newacronym{dvrk}{dVRK}{da Vinci Research Kit}
\newacronym{dagger}{DAgger}{Dataset Aggregation}
\newacronym{cvs}{CVS}{critical view of safety}
\newacronym{fps}{FPS}{frames per second}

\lstdefinestyle{pythonstyle}{
    language=Python,
    basicstyle=\ttfamily\small,
    keywordstyle=\color{blue},
    commentstyle=\color{green!50!black},
    stringstyle=\color{red},
    showstringspaces=false,
    numbers=left,
    numberstyle=\tiny\color{gray},
    frame=single,
    breaklines=true,
    tabsize=4,
}

\tcbset {
  base/.style={
    arc=0mm, 
    bottomtitle=-0.25mm,
    boxrule=0mm,
    colbacktitle=black!10!white, 
    coltitle=black, 
    fonttitle=\bfseries, 
    left=2.5mm,
    leftrule=1mm,
    right=3.5mm,
    title={#1},
    toptitle=0.25mm,
    breakable,
  }
}

\definecolor{brandblue}{rgb}{0.34, 0.7, 1}
\newtcolorbox{mybox}[1]{
  colframe=brandblue, 
  base={#1}
}

\definecolor{pink}{rgb}{1, 0.75, 0.8}
\newtcolorbox{safetybox}[1]{
  colframe=pink, 
  base={#1}
}

\title{Generalized Recognition of Basic Surgical Actions Enables Skill Assessment and Vision-Language-Model-based Surgical Planning}

\correspondingauthor{Yueming Jin (ymjin@nus.edu.sg), Qi Dou (qidou@cuhk.edu.hk)}

\author[1]{\textbf{Mengya Xu}}
\author[2]{\textbf{Daiyun Shen}}
\author[1]{\textbf{Jie Zhang}}
\author[3]{\textbf{Hon Chi Yip}}
\author[4,5]{\textbf{Yujia Gao}}
\author[6]{\textbf{Cheng Chen}}
\author[1]{\textbf{Dillan Imans}}
\author[1]{\textbf{Yonghao Long}}
\author[1]{\textbf{Yiru Ye}}

\author[7]{\textbf{Yixiao Liu}}  
\author[8]{\textbf{Rongyun Mai}} 
\author[1]{\textbf{Kai Chen}}   

\author[9]{\textbf{Hongliang Ren}}
\author[10]{\textbf{Yutong Ban}}  
\author[11]{\textbf{Guangsuo Wang}} 
\author[12]{\textbf{Francis Wong}} 
\author[12]{\textbf{Chi-Fai Ng}} 

\author[13]{\textbf{Kee Yuan Ngiam}} 
\author[14]{\textbf{Russell H. Taylor}} 
\author[15]{\textbf{Daguang Xu}} 
\author[2,16,$\ast$]{\textbf{Yueming Jin}}
\author[1,$\ast$]{\textbf{Qi Dou}}

\affil[1]{Department of Computer Science and Engineering, The Chinese University of Hong Kong, Hong Kong, China}
\affil[2]{Department of Biomedical Engineering, National University of Singapore, Singapore}
\affil[3]{Department of Surgery, The Chinese University of Hong Kong, Hong Kong, China}
\affil[4]{Division of Hepatobiliary and Pancreatic Surgery, Department of Surgery, National University Hospital, Singapore}
\affil[5]{iHealthTech, National University Singapore, Singapore}
\affil[6]{Department of Electrical and Electronic Engineering, The University of Hong Kong, Hong Kong, China}
\affil[7]{Department of Urology, Peking University Third Hospital, Beijing, China}
\affil[8]{Department of Hepatobiliary and Pancreatic Surgery, Guangxi Medical University Cancer Hospital, Guangxi Medical University, Nanning, China}
\affil[9]{Department of Electronic Engineering, The Chinese University of Hong Kong, Hong Kong, China}
\affil[10]{Global College, Shanghai Jiao Tong University, Shanghai, China}

\affil[11]{Department of Thoracic Surgery, Shenzhen People's Hospital (The Second Clinical Medical College, Jinan University; The First Affiliated Hospital, Southern University of Science and Technology), Shenzhen, Guangdong, China}
\affil[12]{Division of Urology, Department of Surgery, The Chinese University of Hong Kong, Hong Kong, China}
\affil[13]{Division of General Surgery (Endocrine and Thyroid Surgery), Department of Surgery, National University Hospital, Singapore}

\affil[14]{Johns Hopkins University, Baltimore, United States}
\affil[15]{NVIDIA Corporation}
\affil[16]{Department of Electrical and Computer Engineering, National University of Singapore, Singapore}

\begin{document}

\begin{abstract}
Artificial intelligence, imaging, and large language models have the potential to transform surgical practice, training, and automation. Understanding and modeling of basic surgical actions (BSA), the fundamental unit of operation in any surgery, is important to drive the evolution of this field. In this paper, we present a BSA dataset comprising 10 basic actions across 6 surgical specialties with over 11,000 video clips, which is the largest to date. Based on the BSA dataset, we developed a new foundation model that conducts general-purpose recognition of basic actions. Our approach demonstrates robust cross-specialist performance in experiments validated on datasets from different procedural types and various body parts. Furthermore, we demonstrate downstream applications enabled by the BAS foundation model through surgical skill assessment in prostatectomy using domain-specific knowledge, and action planning in cholecystectomy and nephrectomy using large vision-language models. Multinational surgeons' evaluation of the language model's output of the action planning explainable texts demonstrated clinical relevance. These findings indicate that basic surgical actions can be robustly recognized across scenarios, and an accurate BSA understanding model can essentially facilitate complex applications and speed up the realization of surgical superintelligence.
\end{abstract}

\maketitle

\section*{INTRODUCTION}
Artificial intelligence (AI) and large language models (LLM) in healthcare are experiencing explosive growth. Their applications in surgery hold substantial potential for enhancing procedural consistency, facilitating skill assessment~\cite{kiyasseh2023vision, cui2024capturing, chen2022surgesture, olsen2024surgical} and learning~\cite{ma2021novel}. Despite recent progress in AI-based surgical video analysis~\cite{schmidgall2024general, yuan2025learning, wei2025surgbench}, current solutions still lack generalizability in practice. Surgery represents a complex interplay of skills and actions. However, this complexity can be decomposed into fundamental components known as Basic Surgical Actions (BSAs). These BSAs present consistent patterns across surgical specialties. As standardized building blocks, these BSAs can be combined in various sequences to generate diverse surgical procedures. While different surgical specialties may exhibit distinct anatomical features and contexts, the underlying BSAs often share common patterns in instrument handling and tissue manipulation. Therefore, developing a systematic understanding of these BSAs is feasible and crucial for advancing surgical data science.

\begin{figure*}[t!]
\centering
\includegraphics[width=1\linewidth]{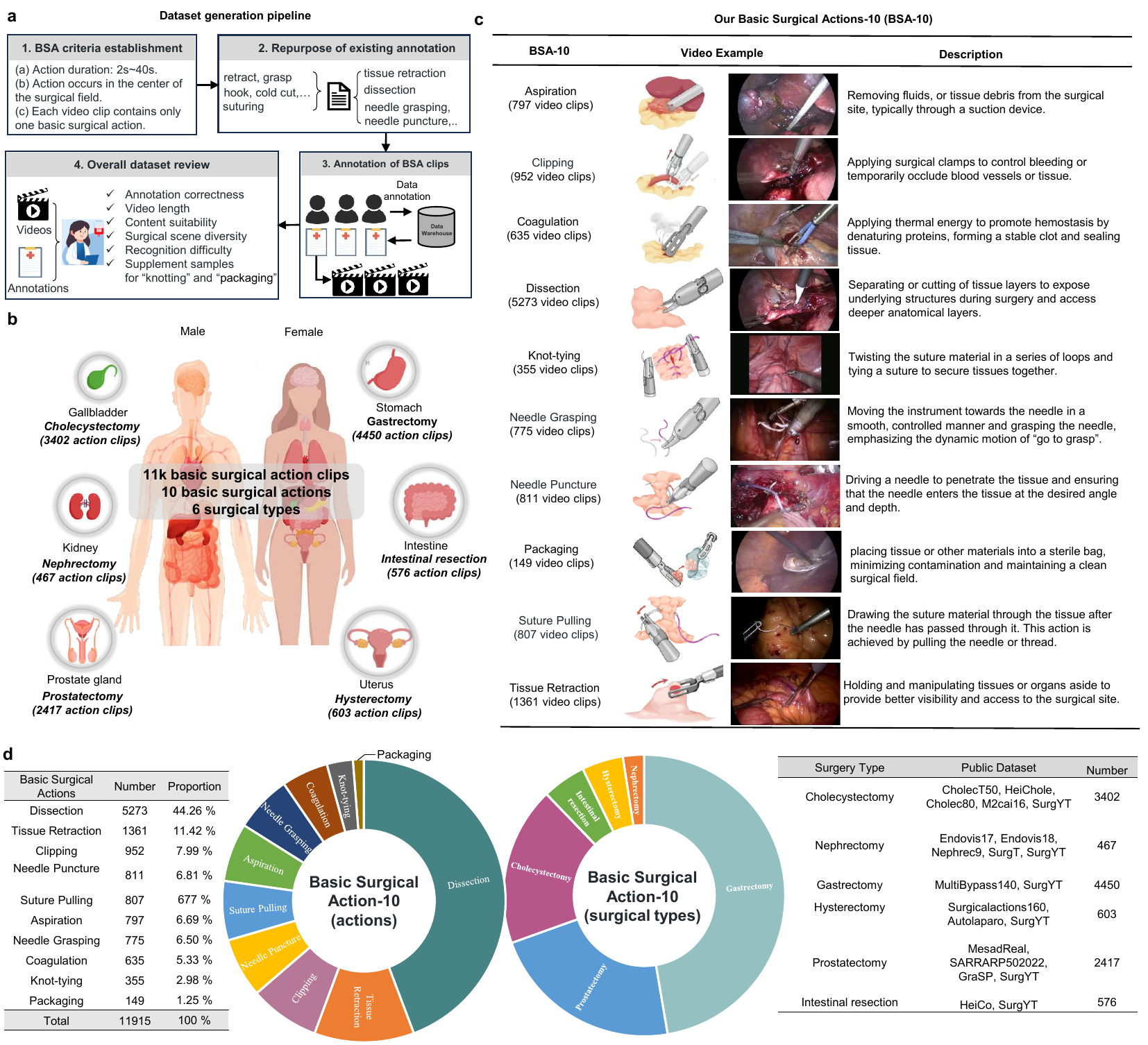}
\caption{{\bf Illustration of our BSA-10 dataset.} \textbf{a} Dataset generation pipeline. The process involves four key steps: BSA criteria establishment, reuse of existing annotations, annotation of new BSA clips, and overall dataset review; \textbf{b} Our dataset is collected from $6$ body parts (i.e., gallbladder, stomach, kidney, intestine, prostate gland, and uterus) and sourced from $15$ public datasets and our SurgYT collection (surgical videos from credentialed YouTube channels); \textbf{c} Description of basic surgical actions; \textbf{d} Statistical analysis of the dataset across different basic surgical actions and procedure types.}
\label{fig_meta_action_dataset}
\end{figure*}

Previous work has investigated a similar concept of basic surgical activities and reported promising findings~\cite{hung2017structured}. Kiyasseh et al.~\cite{kiyasseh2023vision} proposed a vision transformer to decode three needle-based surgical activities and showed its value for skill analysis in different urology procedures. Despite its limited scope of procedural types, their study yielded insights that understanding these basic actions can influence surgical behaviors and correlate with patient outcomes. This suggests that expanding the study of Basic Surgical Actions by increasing the number of BSAs and types of procedures could have far-reaching implications.

Recent foundation models~\cite{schmidgall2024general, wei2025surgbench} and LLM technologies~\cite{singhal2025toward, wang2025endochat, low2025surgraw} offer general-purpose recognition and reasoning capabilities. However, existing surgical datasets~\cite{chen2022surgesture, ahmidi2017dataset, gao2014jhu, shafiei2021surgical, kitaguchi2020automated}, limited by their size and procedural scope, cannot fully unleash these powerful computational tools. This limitation underscores the critical need for comprehensive, large-scale datasets to facilitate research in surgical AI. This universality in basic actions not only facilitates activity recognition but also enables various downstream applications. These applications include automated surgical step identification~\cite{hashemi2025video} and objective skill evaluation~\cite{sirajudeen2024deep}. Furthermore, understanding BSAs allows for the development of long-horizon surgical planning systems, potentially advancing surgical automation through a hierarchical approach to procedural decomposition.

In this study, we present a novel approach to surgical action analysis through the development of the largest and most comprehensive Basic Surgical Action (BSA) dataset. This dataset encompasses 10 basic surgical actions across 6 surgical specialties, comprising over 11k video clips. We introduce a foundation model for generalized BSA recognition and validate its performance on two external datasets representing different procedural types. Additionally, we demonstrate the model's utility in downstream applications, including surgical skill assessment in prostatectomy procedures and surgical action planning in cholecystectomy and nephrectomy, utilizing multimodal LLM integration. The clinical relevance of our findings is evaluated by a multi-national panel of surgeons from diverse healthcare systems, providing expert validation and a globally representative assessment of the system's outputs.

\section*{\textbf{RESULTS}}
\subsection*{A New Dataset of Basic Surgical Actions}
We built a comprehensive dataset of basic surgical actions involving 10 actions that commonly exist in laparoscopic and robotic surgery (named BSA-10 dataset). Figure~1 shows an overview of the BSA-10 dataset, with basic actions including aspiration, clipping, coagulation, dissection, knot-tying, needle grasping, needle puncture, packaging, suture pulling, and tissue retraction. The dataset has a total of \textcolor{blue}{$11,915$} video clips covering different types of procedures in 6 body parts (i.e., gallbladder, stomach, kidney, intestine, prostate gland, and uterus). The video clips were sourced from $15$ public surgical datasets and our SurgYT collection (surgical videos from credentialed YouTube channels). The public datasets included CholecT50~\cite{nwoye2022rendezvous}, Cholec80~\cite{twinanda2016endonet}, M2cai16-workflow~\cite{cadene2016m2cai}, Heichole~\cite{wagner2023comparative}, MultiBypass140~\cite{ramesh2023weakly}, Surgicalactions160~\cite{schoeffmann2018video}, Autolaparo~\cite{wang2022autolaparo}, HeiCo~\cite{maier2021heidelberg}, MesadReal~\cite{bawa2021saras}, SAR-RARP50 2022~\cite{psychogyios2023sar}, Endovis17~\cite{allan20202018}, Endovis18~\cite{allan20202018}, Nephrec9~\cite{penza2018endoabs}, GraSP~\cite{ayobi2024pixel}, SurgT~\cite{cartucho2024surgt}.

The ultimate dataset consisted of 11,915 surgical video clips covering 10 basic surgical action classes: 797 aspiration; 952 clipping; 635 coagulation; 5,273 dissection; 355 knot-tying; 775 needle grasping; 811 needle puncture; 149 packaging; 807 suture pulling; and 1,361 tissue retraction. These clips spanned six surgical procedures: 3,402 cholecystectomy clips; 4,450 gastrectomy clips; 603 hysterectomy clips; 576 intestinal resection clips; 467 nephrectomy clips; and 2,417 prostatectomy clips.

% Laparoscopic Roux-En-Y Gastric Bypass (LRYGB)~\cite{leyba2014laparoscopic}, 

To construct the dataset that focuses on BSA, we repurposed the public data following a systematic protocol (Figure~\ref{fig_meta_action_dataset}a). Briefly speaking, we followed four steps:
\textit{1) BSA criteria establishment:} specifying the definition of basic actions and developing the criteria of high-quality video clips;
\textit{2) Repurpose of existing annotations:} identifying and utilizing video clips from current datasets where their existing labels aligned with our BSA protocol, particularly when their annotated surgical actions matched our BSA definitions;
\textit{3) Annotation of new BSA clips:} annotating new video clips from the surgical videos that did not address BSA annotations, thereby repurposing the public data by extracting BSA clips;
\textit{4) Overall dataset review:} quality control through expert review and data cleaning. 
Statistics of the BSA-10 dataset, including the number of video clips for each basic surgical action, are shown in Figure~\ref {fig_meta_action_dataset}(c-d). More details about dataset construction are provided in Sec. Materials and Methods.

\subsection*{Generalized BSA Recognition Model on BSA-10 Dataset}
Based on the BSA-10 dataset, we developed a BSA foundation model, a transformer-based network designed for generalized recognition of the basic surgical actions. 
Using one single model to simultaneously recognize ten different basic surgical actions across many distinct procedures is challenging. To achieve this, we designed a new transformer-based architecture which can capture highly representative spatio-temporal information to distinguish visual appearances and motion patterns. Detailed methods are described in Sec. Materials and Methods.

We conducted ten-fold cross-validation for the BSA recognition model. To evaluate the performance of the foundation model, we computed the receiver operating characteristic (ROC) curve and measured the area under the ROC curve (AUROC) using the one-versus-all approach~\cite{aspart2022clipassistnet}. 
We also calculated the sensitivity and specificity using a threshold determined by the Youden Index~\cite{ruopp2008youden} derived from the ROC curve. 
To analyze the model's robustness across different surgical actions and anatomical regions, we observed the averaged evaluation metrics for both basic actions and body parts.

\begin{figure*}[t!]
\centering
\includegraphics[width=1\linewidth]{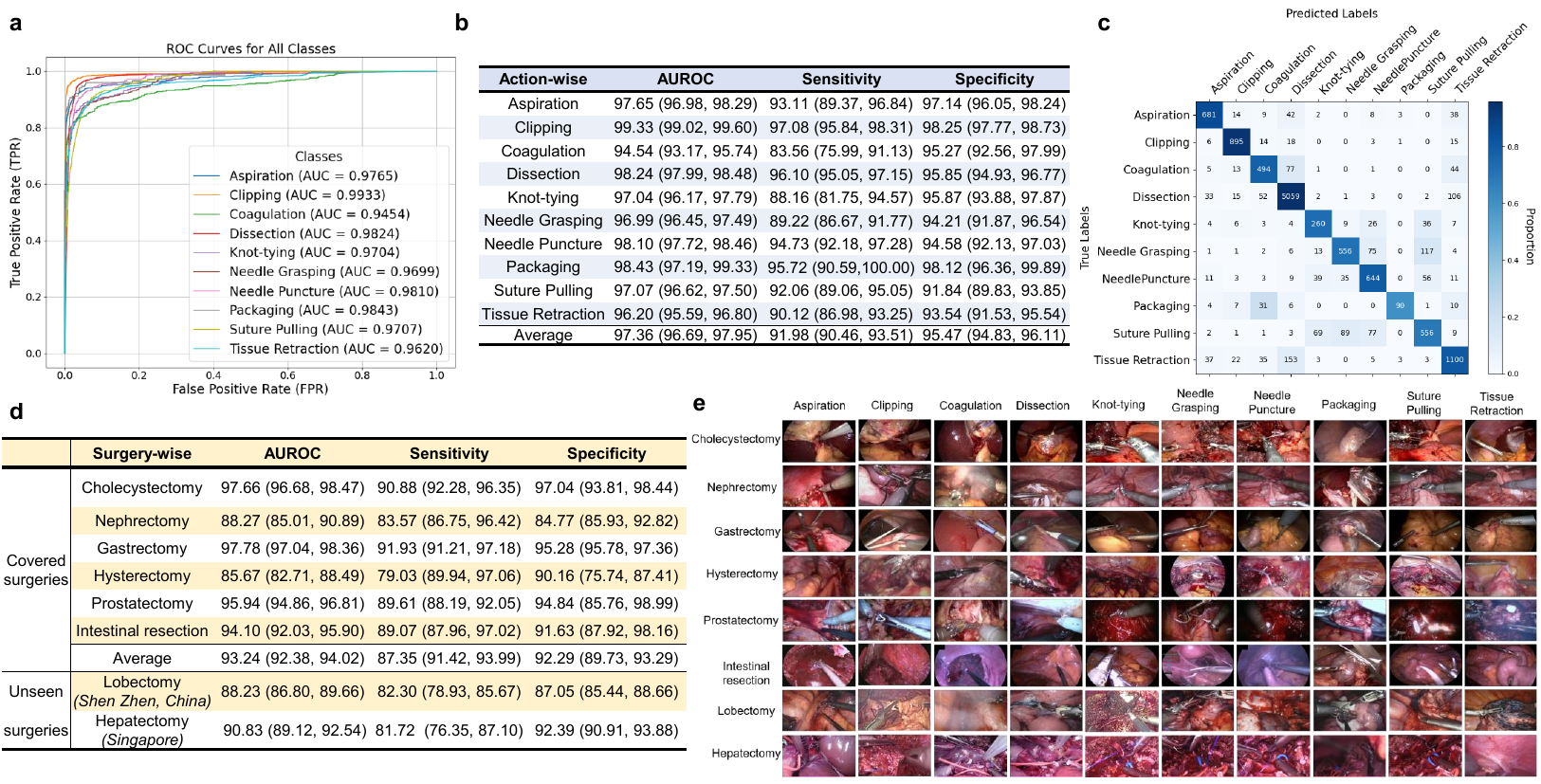}
\caption{{\bf Analysis results of 10-fold cross-validation on the developed dataset.} 
\textbf{a} The receiver operating characteristic (ROC) curve of ten action classes;
\textbf{b} Results of the model on ten action classes based on the Youden Index which are presented as 95\% confidence interval; 
\textbf{c} The confusion matrix across ten action classes by aggregating the individual confusion matrices from each of the ten folds; 
\textbf{d} Surgery-wise performance metrics, displayed with 95\% confidence intervals;
\textbf{e} Representative frames from ten basic surgical actions across eight surgery types, highlighting intra-class variability.
}
\label{recognition_results}
\end{figure*}

\noindent
\textbf{Performance of BSA recognition across different actions and surgical types:}
~Figure~\ref{recognition_results} (a-b) presents the action-wise results of the BSA recognition model. All ten surgical actions received recognition with AUROC scores above 0.9. The overall average sensitivity was 91.98\% (CI: 90.46\%, 93.51\%) and average specificity was 95.47\% (CI: 94.83\%, 96.11\%). These results supported our hypothesis that basic surgical actions exhibit consistent and identifiable patterns across different surgical specialties. Among the basic surgical actions, packaging and vessel clipping showed the strongest performance. Packaging achieved an AUROC of 98.43\% (CI: 97.19\%, 99.33\%), sensitivity of 95.72\% (CI: 90.59\%,100.00\%), specificity of 98.12\% (CI: 96.36\%, 99.89\%). Clipping achieved an AUROC of 99.33\% (CI: 99.02\%, 99.60\%), sensitivity of 97.08\% (CI: 95.84\%, 98.31\%), specificity of 98.25\% (CI: 97.77\%, 98.73\%). This superior performance could be attributed to the distinctive visual features associated with these actions, which involve specific surgical instruments such as bags for packaging and clamps for vessel clipping. Other actions exhibited inter-class similarity in appearances thus were relatively more challenging to distinguish (see Figure~\ref{recognition_results}c for the confusion matrix). For instance, the BSA recognition model had to capture spatio-temporal representations to accurately classify the needle-related actions, including needle grasping (AUROC: 96.99\%), needle puncture (AUROC: 98.10\%), suturing pulling (AUROC: 97.07\%), and knot-tying (AUROC: 97.04\%). It was also observed that tissue retraction was occasionally misclassified into dissection, as they both generated tissue deformations in the surgical scenes, leading to a peak in the confusion matrix.

For specialist-wise performance (see Figure~\ref{recognition_results}d), the overall average AUROC for the different surgical specialists is 93.24\% (CI: 92.38\%, 94.02\%), demonstrating the general efficacy of the learned network for recognizing basic surgical actions. Cholecystectomy ($\pm$ bile duct exploration) obtained the highest performance with an AUROC of 97.66\% (CI: 96.68\%, 98.47\%), reflecting the standardization of laparoscopic gallbladder removal and the abundant training data available for this procedure. 
Hysterectomy presents the most challenging scenario with an AUROC of 85.67\% (CI: 82.71\%, 88.49\%), due to the anatomical variations and diverse approaches. Intermediate performance is observed in nephrectomy (88.27\%, CI: 85.01\%, 90.89\%) and Intestinal resection (94.10\%, CI: 92.03\%, 95.90\%), reflecting moderate anatomical complexity and procedural variability. Gastrectomy (97.78\%, CI: 97.04\%, 98.36\%) and prostatectomy (95.94\%, CI: 94.86\%, 96.81\%) demonstrate good performance, indicating effective adaptation to specialized anatomical contexts while maintaining BSA recognition accuracy.

\subsection*{Evaluation on Multi-national External Data of Unseen Surgical Types}

To further evaluate the generalizability of our BSA recognition model on unseen surgical types, we conducted external validation on procedures that were not included in the BSA-10 training dataset.
We first evaluated our model on lobectomy data collected from the Department of Thoracic Surgery, Shenzhen People's Hospital, Shenzhen, China, comprising 331 action clips from 16 patients. 
Thoracic surgery in the lung area presents characteristic anatomical appearances and instrumentation. 
For instance, lung parenchyma exhibits spongy, air-filled architecture with fragile pleural surfaces. Thoracic-specific surgical tools include endostaplers, atraumatic lung graspers, and specialized retractors designed for confined thoracic spaces.
Given these challenges, the BSA foundation model still demonstrated strong generalizability, achieving an AUROC of 88.23\% (CI: 86.80\%, 89.66\%), sensitivity of 82.30\% (CI: 78.93\%, 85.67\%), and specificity of 87.05\% (CI: 85.44\%, 88.66\%). 

Among these basic actions, our BSA recognition model achieved excellent performance for packaging with an AUROC of 93.77\% (CI: 89.45\%, 98.09\%), despite the external data's specimen bags having different opacity characteristics compared to the transparent bags in training data. Clipping demonstrated good performance (AUROC: 89.63\%, CI: 85.32\%, 93.93\%) despite differences in clamp types, with lobectomy employing titanium clips for major pulmonary vessels contrasting with the mixed titanium and polymer clips used in other procedures within the training dataset. However, certain actions presented greater challenges in the thoracic context. Coagulation achieved moderate performance with an AUROC of 79.11\% (CI: 75.14\%, 83.09\%), reflecting the specialized energy delivery patterns required for lung parenchyma that differ from standard electrocautery techniques. The detailed performance metrics for lobectomy procedures across all ten BSA classes are presented in the supplementary materials Table~\ref{tab_supp_1}.

Considering that surgical techniques and skills may vary across different surgical practices and geographical regions, we further tested the BSA model on external data not only from an unseen surgical type, but also from a different country. Specifically, we collected the hepatectomy videos from the Department of Surgery, National University Hospital, Singapore, consisting of 210 action clips from 6 patients.
Our model achieved an AUROC of 90.83\% (CI: 89.12\%,92.54\%), sensitivity of 81.72\% (CI: 76.35\%, 87.10\%), and specificity of 92.39\% (CI: 90.91\%, 93.88\%). While the performance was relatively lower, this degradation reflects the combined challenges of the novel procedure and potential variations in surgical skills across different geographic regions. Supplementary Table~\ref{tab_supp_2} presents the detailed performance metrics for hepatectomy procedures across all ten BSA classes.
Figure~\ref{recognition_results}e illustrates examples from ten BSAs in eight specialized surgeries, showing the diversity in anatomical regions, surgical tools, procedural techniques, and tissue characteristics.
These procedure-specific variations underscore the model's ability to maintain consistent BSA recognition across diverse surgical contexts.
The generalization to both unseen procedures (lobectomy and hepatectomy) and multi-national data (China and Singapore) indicates good potential for the wide applicability of the BSA recognition model.

\subsection*{BSA Recognition Facilitates Surgical Skill Assessment}

Operative skill analysis is essential in surgical education and practice. Skill assessment can streamline evaluation, reduce manual effort, and eliminate subjectivity. Additionally, modeling optimal surgical execution can guide the design of control policies, enabling robotic systems to mimic expert surgeon performance. 

\begin{figure*}
\centering
\includegraphics[width=1\linewidth]{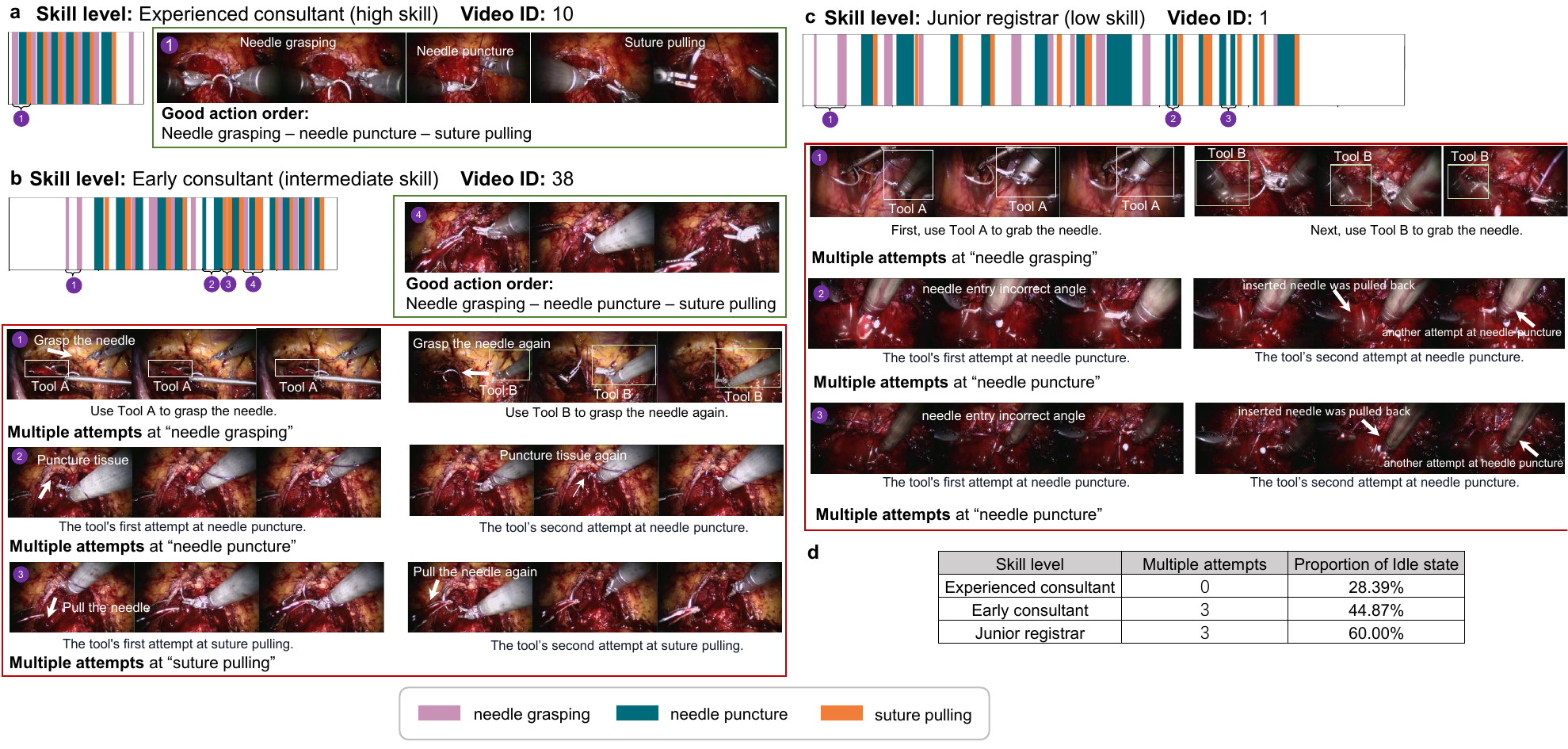}
\caption{{\bf Action barcode visualization of BSA distributions across expertise levels}. The skill analysis for the three surgical procedures conducted by the \textbf{a} experienced consultant, \textbf{b} the early consultant, and \textbf{c} the junior registrar in RARP50. The action barcode delineates three BSAs, namely needle grasping, needle puncture, and suture pulling, represented by distinct colors: purple, blue, and orange, respectively. \textbf{d} Quantitative comparison of multiple-attempt frequency and idle-state proportion across three surgical expertise levels.}
\label{skill}
\end{figure*}

There exist high correlations between skill performance and the executed action during the surgery. In this regard, our BSA recognition model can facilitate skill analysis by generating action barcodes, as shown in Figure~\ref{skill}. An action barcode represents the surgical procedure as color-coded temporal action sequences, where each BSA is assigned a distinct color and the duration of each action corresponds to the width of its respective segment. Through analysis of these action barcodes, we can identify the following two factors that highly correlate with expertise skill levels: (i) \textit{The frequency of multiple attempts:} When surgeons lack proficiency, they tend to repeat the same action multiple times before successfully transitioning to the next procedural step. This repetitive behavior disrupts the expected sequential action order and manifests as clusters of identical actions in the temporal sequence.  For example, repeated needle grasping attempts before achieving proper needle puncture, or multiple needle puncture actions before successful tissue penetration. (ii) \textit{Proportion of non-action state:} The rate of idle states (non-action), calculated by subtracting the total action duration from the overall procedure length, reflects the proportion of surgeons' hesitations and proficiency. 

We employ the SAR-RARP50 dataset~\cite{sirajudeen2024deep}, which includes 50 in-vivo suturing segments extracted from Robotic Assisted Radical Prostatectomy (RARP) procedures,  to demonstrate the effectiveness of our action recognition method for skill analysis. The RARP50 dataset includes three-level skills and three-class actions. Skills are categorized by expertise level from high to low as: experienced consultant (high skill, $23$ surgical videos), early consultant (intermediate skill, $14$ surgical videos), and junior registrar (low skill, $3$ surgical videos). The SAR-RARP50 video dataset captures suturing of the dorsal vascular complex (DVC), a network of veins and arteries that is sutured to minimize bleeding following transection of the prostate's connections to the bladder and urethra~\cite{sirajudeen2024deep}. Therefore, actions include needle grasping, needle puncture, and suture pulling. Since the SAR-RARP50 dataset is a subset of the BSA dataset, we evaluate prostatectomy procedures using a 10-fold cross-validation strategy, where each fold's trained model generates predictions on its corresponding test set. Predictions from all individual folds are aggregated to produce the final BSA action recognition results. The overall performance is reported as the average accuracy across all folds, achieving a final action recognition accuracy of 83.32\%.

As depicted in Figure~\ref{skill}a, the \textbf{experienced consultant} demonstrates exemplary technical proficiency with a well-organized action sequence, indicated by zero multiple attempts and an efficient procedure where the idle state accounts for only 28.39\%  over a total procedure
time of 2min 7s. This performance reflects expert-level procedural planning and execution confidence, with minimal hesitation or multiple attempts required. 
The \textbf{junior registrar} demonstrates the poorest performance, with multiple attempts during both needle grasping and needle puncture actions, accompanied by a substantially elevated idle state proportion of 60\% (see Figure~\ref{skill}c) over a prolonged duration of 9min 22s. Detailed examination reveals underlying performance mechanisms: incorrect needle grasping (at the tail rather than the body) prevents direct progression to puncture, necessitating multiple re-grasping attempts. Subsequent idle periods represent cognitive processing time for error recognition and corrective strategy formulation. Further analysis reveals that the surgeon subsequently struggles with incorrect needle positioning, requiring multiple needle puncture attempts to achieve proper tissue penetration. 
The \textbf{early consultant} exhibits intermediate performance patterns with $3$ multiple attempts and relatively shorter idle state rates of 44.87\%, completed within a
duration of 5min 5s (Figure~\ref{skill}b). The surgeon eventually adheres to the correct action sequence (needle grasping, needle puncture, and suture pulling), demonstrating the learning capacity characteristic of intermediate expertise. This pattern reflects the transitional nature of intermediate expertise, where procedural knowledge enables rapid error correction even when initial execution requires multiple attempts. These BSA-derived insights provide the interpretable characterization of skill analysis, highlighting their potential as a tool for objective surgical skill assessment.

\subsection*{BSA for Surgical Action Planning with Multimodal Large Language Model}

\begin{figure*}
\centering
\includegraphics[width=1\linewidth]{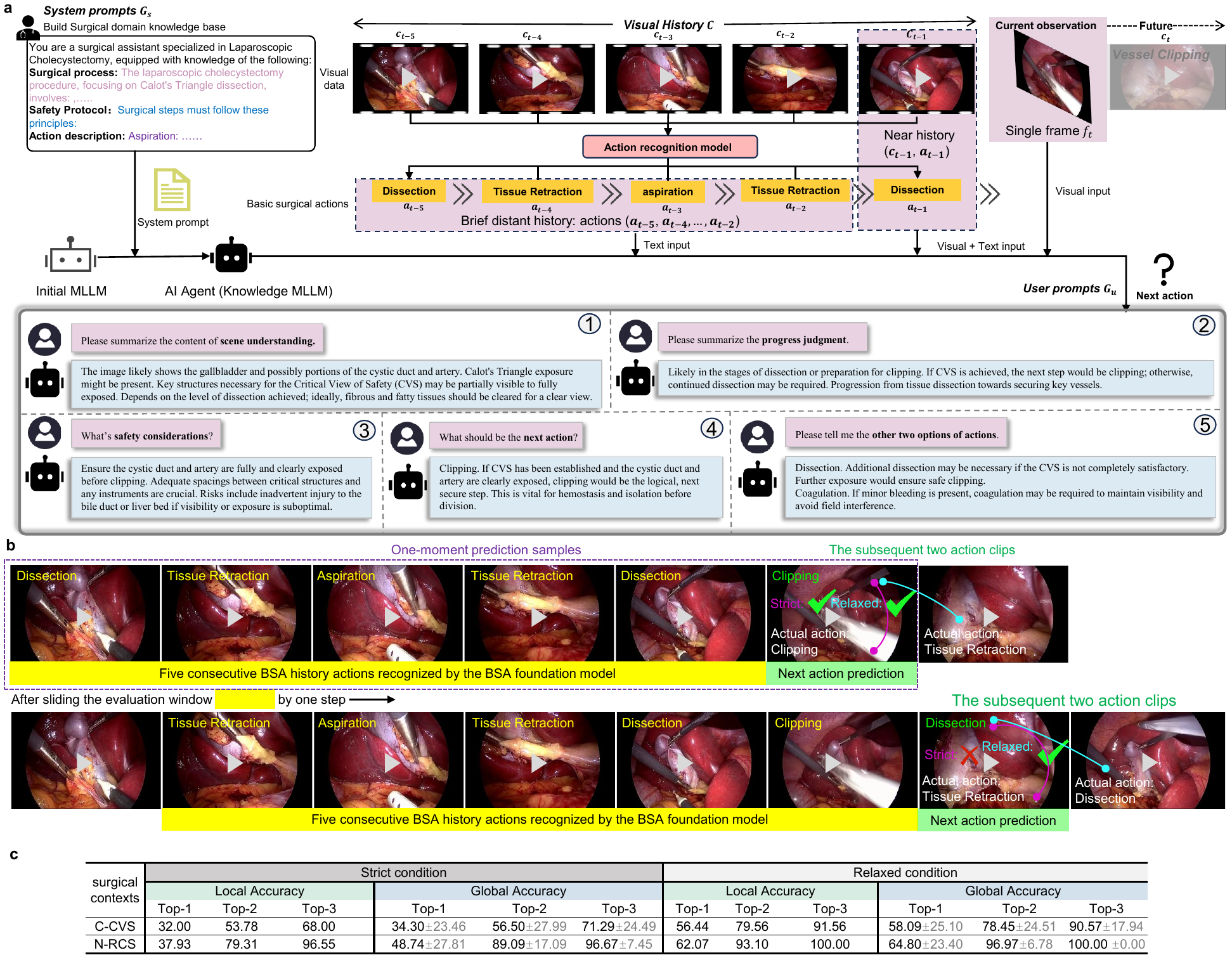}
\caption{{\bf BSA for surgical action planning with our AI agent.} \textbf{a} Our AI Agent integrates surgical context knowledge, historical data, and current observations as inputs, generating a response upon receiving a user prompt. 
\textbf{b} Our AI Agent's reasoning process through two representative examples from the C-CVS scenario.
\textbf{c} Next action prediction results. We report the local and global accuracy under both strict and relaxed conditions for the C-CVS and N-RCS scenarios.
}
\label{planning_results}
\end{figure*}

A key application of basic surgical action recognition is to use these fundamental units to model extended action sequences over a longer period. This enables surgical action planning for more complex contexts. By continuously monitoring previous actions, the AI agent can automatically predict the next action or sequence of actions. We formulated such a surgical action planning problem in a way that can leverage the reasoning ability of advanced vision language models (VLM). 
Figure~\ref{planning_results} shows the idea. Specifically, our VLM-based AI agent relies on GPT-4o~\cite{achiam2023gpt} with domain-specific prompt engineering. First, we provide GPT-4o with a system prompt including texts about the surgical process, safety protocol, and basic action descriptions. This system prompt provides essential background information to understand the specified surgical context. Second, we use temporally stratified procedural history comprising distant history (recognized 4 BSA clips from the earlier procedural process) and recent history (the most recent action clip with its corresponding recognized action), Third, we provide these inputs to GPT-4o, then ask GPT-4o to predict the most possible next action, for at most three possibilities. When doing the prediction, we prompted the GPT-4o to consider aspects of scene understanding, progress judgment, and safety considerations.

\noindent
\textbf{Metrics definition}
To evaluate the performance of the surgical action planning task in predicting the next basic action, we used top-1, top-2, and top-3 prediction accuracy metrics within a comprehensive evaluation framework. Each surgical video context contains a sequence of BSA clips. The AI agent takes five consecutive BSA clips as input to predict the next action, with this one-moment prediction accuracy termed as \textit{``local accuracy''}. 
Then, the evaluation window slides forward, predicting each subsequent basic action until reaching the context's end. We calculate \textit{``global accuracy''} as the average of all predictions within a context, measuring the model's consistent performance across the entire long period rather than isolated moments.
Additionally, given that surgical actions can often be performed in a flexible order (i.e., some actions can be performed earlier or later without affecting the procedure's outcome)~\cite{standring2005gray,strasberg2017critical,padoy2012statistical}, we also introduced a ``relaxed metric'' that considers predictions correct if they appear within the next two actions of the ground truth sequence. More details can be found in Sec. Materials and Methods.

\noindent
\textbf{Performance of surgical action planning with GPT-4o}
We experimented on two scenarios demonstrating typical complex surgical contexts that could be decomposed into a sequence of basic surgical actions. 
% The first scenario was extracted from cholecystectomy, and the second scenario was extracted from nephrectomy. 

\noindent\textbf{Scenario-1: Achieving critical view of safety in Cholecystectomy (C-CVS):}
We strategically selected the Critical View of Safety (CVS) achievement process as our BSA-based surgical planning scenario due to its unique clinical significance and technical complexity. The process starts with the attempt to expose the Calot's triangle and ends with the transaction of the cystic duct. CVS~\cite{murali2023latent, mascagni2022artificial, strasberg1995analysis} represents the most safety-critical phase of cholecystectomy, requiring identification and exposure of key anatomical structures before any dissection or clipping. This phase demands precise coordination of five BSAs, including aspiration, coagulation, dissection, tissue retraction, and clipping, to minimize bile duct injury risk. Our extracted 50 C-CVS contexts from the CholecT50 dataset~\cite{nwoye2022rendezvous}, which contains 50 patient videos, span durations from 4.49 to 8.23 minutes (average: 6.36 minutes) and focus on achieving CVS criteria through coordinated BSA execution. From these 50 context segments, we derived 225 one-moment prediction samples as the evaluation window slides forward.

The CholecT50 dataset is encompassed within our BSA dataset. Our BSA recognition model achieved $90.77\%$ accuracy on the cholecystectomy action clips. These context-specific recognition accuracies establish the reliable foundation necessary for surgical action planning, where accurate interpretation of procedural history enables dynamic prediction of subsequent actions.

We illustrate our AI agent's reasoning through two examples (Figure~\ref{planning_results}b). In the first, the agent processes five BSA-recognized actions (dissection, retraction, aspiration, retraction, dissection) and predicts clipping, reasoning: ``CVS is established; clipping is the logical next step.'' The subsequent actions (clipping, retraction) confirm this prediction, contributing to 68.00\% strict and 91.56\% relaxed top-3 accuracy. In the next window (updated actions: retraction, aspiration, retraction, dissection, clipping), the agent predicts dissection, reasoning: ``Needed to expose the cystic duct.'' The ground truth (retraction, dissection) matches only under relaxed conditions (56.44\% top-1 accuracy), reflecting surgical workflow variability. This process iterates across all 50 contexts, with the evaluation window sliding forward to predict each subsequent basic action until reaching the context's end. As shown in Figure~\ref{planning_results}c, while strict next-action prediction remains challenging (top-1 global accuracy: 34.00\%), the dramatic improvement to top-3 global accuracy (71.29\%) indicates that clinically appropriate actions consistently appear within high-confidence recommendations. Under relaxed evaluation, our top-3 local and global accuracies of 91.56\% and 90.58\% demonstrate that valid surgical decisions reliably emerge from the model's reasoning process.

\textbf{Scenario-2: Nephrectomy renal cortex suture (N-RCS):} 
We selected renal cortex suturing during robot-assisted laparoscopic nephrectomy as our second planning scenario~\cite{gettman2004robotic, nakawala2019deep}. This task is a short yet technically demanding micro-context that requires the surgeon to execute three basic surgical actions in strict order: needle grasping, needle puncture, and suture pulling. Coordination of BSAs is essential to approximate the renal parenchyma while avoiding additional tissue trauma and preserving residual renal function. We collected 10 videos of nephrectomy procedures from Prince of Wales Hospital (Hong Kong), a dataset independent of the BSA dataset. We extracted 10 renal-suturing segments from these videos, each beginning with a needle grasping and ending with a suture pulling for tightening. These representative context segments have durations ranging from 48 to 84 seconds (average: 63.6 seconds). From these 10 carefully selected context segments, we derived 29 one-moment prediction samples as the evaluation window slides forward.

The BSA recognition model achieved $62.10\%$ on nephrectomy action clips. The AI Agent processes five consecutive BSA history actions  (e.g., suture pulling, suture pulling, needle puncture, suture pulling, needle puncture) and predicts the next action as suture pulling. The actual two subsequent actions are suture pulling and needle puncture. This prediction is considered correct under both strict and relaxed conditions. As the evaluation window progresses through all the complete surgical contexts, our AI Agent achieves the top-1 \textit{local accuracy} of 37.93\% and \textit{global accuracy} of 48.74\% under the strict condition. Under the relaxed conditions, the model obtains top-1, top-2, and top-3 \textit{local accuracy} of $62.07\%$, $93.10\%$, and $100.00\%$, respectively. The top-1, top-2, and top-3 \textit{global accuracy} are $64.80\%$, $96.97\%$ $100.00\%$. This performance demonstrates the model's ability to provide clinically relevant recommendations, supporting its potential for real-world surgical decision-making assistance.

These findings validate our framework's scalability across diverse procedural scenarios. Collectively, these two scenarios (C-CVS and N-RCS) encompass 8 out of our 10 defined BSAs, demonstrating comprehensive BSA coverage across diverse procedural and anatomical contexts.

\begin{figure*}[t!]
\centering
\includegraphics[width=1\linewidth]{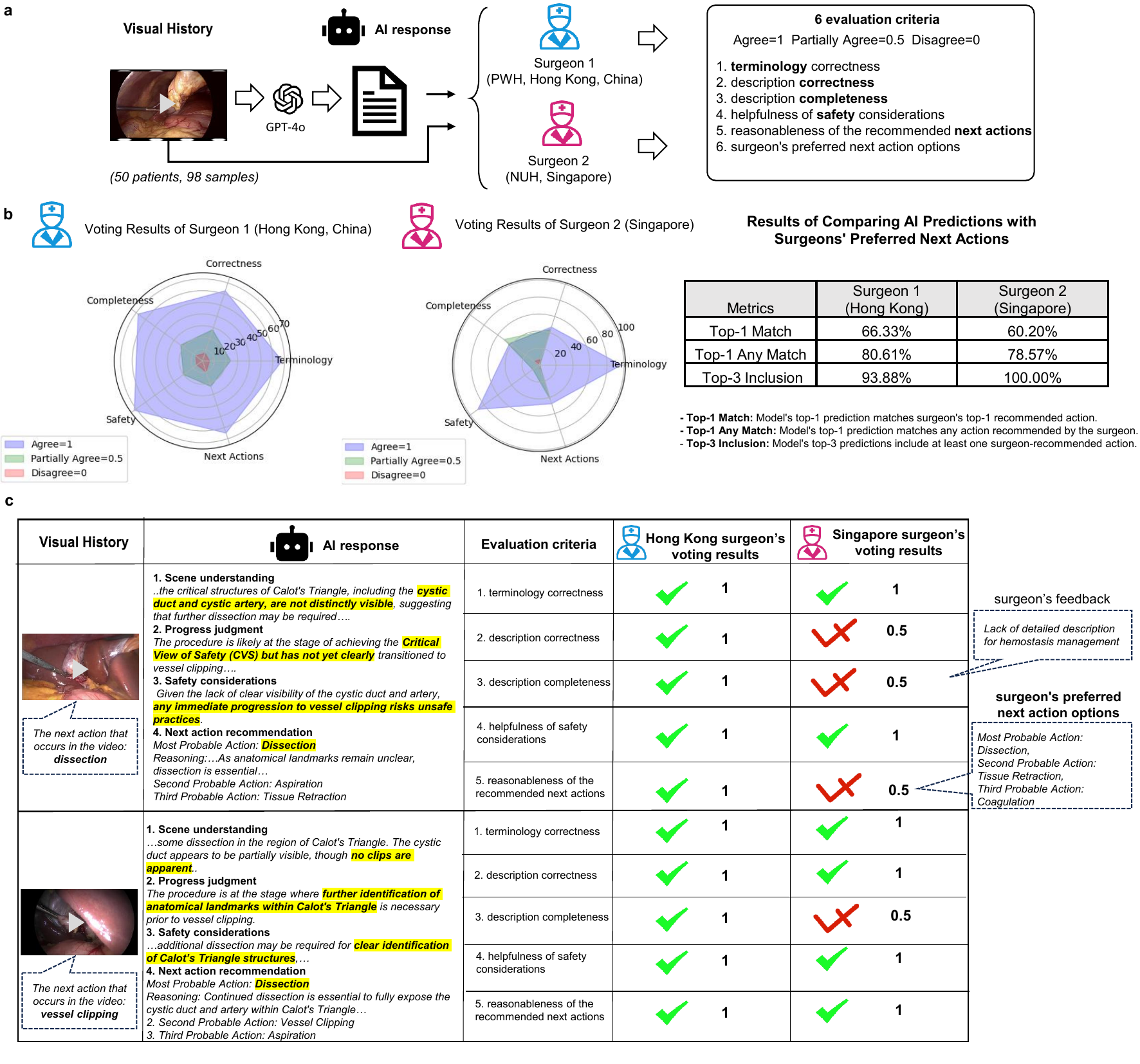}
\caption{{\bf Multi-national surgeon evaluation of AI-generated surgical reasoning and action recommendations.} \textbf{a} Comprehensive scoring evaluation process by surgeons across five clinical criteria, including terminology correctness, description correctness, description completeness, safety considerations, and next action reasonableness. \textbf{b} Quantitative surgeon scoring results from Hong Kong and Singapore surgeons across 98 samples, demonstrating international consensus on AI output quality. \textbf{c} Representative examples of detailed surgeon scoring with specific feedback highlighting clinical judgment variations and the importance of flexible evaluation metrics in surgical AI systems.}
\label{surgeon_evaluation}
\end{figure*}

\subsection*{Human Evaluation from Multi-national Surgeons for Language Outputs}
Since the VLM-based AI agent generates clinical recommendations and intraoperative judgments, it is critical to evaluate the clinical accuracy and reliability of its textual outputs.
To assess the quality of the textual responses generated by the VLM, we conducted a multinational surgeon assessment study involving two surgeons: one from Prince of Wales Hospital (PWH) in Hong Kong and one from National University Hospital (NUH) in Singapore. We randomly selected a subset of $98$ one-moment samples from 50 patient videos for surgeon evaluation and collected their scores.
As shown in Figure~\ref{surgeon_evaluation}a, the surgeon first reviewed the video capturing the visual history for each sample, followed by an examination of the VLM-generated textual response. Subsequently, two surgeons independently cast their votes for the textual response from the following key criteria: \textbf{\textit{1. terminology correctness}} (ensuring the use of accurate and contextually appropriate medical terms); 
\textbf{\textit{2. description correctness}} (verifying the factual accuracy of the information provided); 
\textbf{\textit{3. description completeness}} (assessing whether the response covers all necessary aspects of the query);
\textbf{\textit{4. helpfulness of safety considerations}} (determining the practical utility and clarity of the response about the safety considerations in supporting surgical decision-making);
\textbf{\textit{5. reasonableness of the recommended next actions}} (evaluating whether the three probable actions recommended by the model align with the surgeon's clinical decisions and preferences);
\textbf{\textit{6. surgeon's preferred next action options}}
(Obtaining the three options for the next action from the surgeons).
Agreement scores were assigned as follows: Agree = 1, Partially Agree = 0.5, and Disagree = 0. The experts voted 0.5 for mostly correct texts with minor issues, indicating partial agreement.

The scoring results across 98 samples from two surgeons with multi-national backgrounds are shown in Figure~\ref{surgeon_evaluation}b. Multinational surgeon evaluation demonstrates a generally high level of international consensus on output quality across five evaluation criteria, while revealing distinct patterns in clinical assessment approaches.
For \textit{terminology correctness}, the Singapore surgeon demonstrated exceptional approval (97 agreed, 1 partially agreed, 0 disagreed) compared to the Hong Kong surgeon's more measured assessment (68 agreed, 24 partially agreed, 6 disagreed). 
\textit{Safety considerations} received particularly strong endorsement from two surgeons (72 agreed, 20 partially agreed, 6 disagreed from the Hong Kong surgeon; 89 agreed, 9 partially agreed, 0 disagreed from the Singapore surgeon).
For \textit{description correctness}, the Singapore surgeon exhibited greater scrutiny (47 agreed, 44 partially agreed, 7 disagreed) compared to the Hong Kong surgeon's more favorable evaluation (63 agreed, 28 partially agreed, 7 disagreed). Similarly, \textit{reasonableness of the recommended next actions} revealed the Singapore surgeon's conservative approach (48 agreed, 42 partially agreed, 8 disagreed) compared to the Hong Kong surgeon's evaluation (65 agreed, 23 partially agreed, 10 disagreed), suggesting stricter standards for action recommendation validation.

Given the inherent variability in surgical decision-making patterns and procedural preferences across different clinical practices, we further assess the alignment between AI-predicted next surgical actions and multi-national surgeons' perspectives. We designed three metrics: \textit{Top-1 Match} (Model's top-1 prediction matches surgeon's top-1 recommended action), \textit{Top-1 Any Match} (Model's top-1 prediction matches any action recommended by the surgeon), and  \textit{Top-3 Inclusion} (Model's top-3 predictions include at least one surgeon-recommended action). 
Cross-regional validation showed consistent AI-surgeon alignment (see Figure~\ref{surgeon_evaluation}b), with 66.33\%/80.61\% (Hong Kong) and 60.20\%/78.57\% (Singapore) Top-1 match rates, reflecting institutional training influences. Crucially, Top-3 inclusion reached 93.88\%-100.00\% across regions, demonstrating the AI's robust capture of clinically valid options despite surgical preference variations, a key requirement for decision-support systems.

Exploring the differences in scoring outcomes between surgeons in Singapore and Hong Kong would provide further insights. Figure~\ref{surgeon_evaluation}c illustrates surgeon feedback across two representative samples, revealing distinct evaluation patterns between international assessors. In the first sample, the Hong Kong surgeon fully agreed (score 1.0 across all five criteria), reflecting confidence in the AI's comprehensive analysis. Conversely, the Singapore surgeon assigned partially agree scores (0.5) for \textit{description correctness} and \textit{description completeness}, citing omitted hemorrhage management. Additionally, this surgeon rated \textit{reasonableness of the recommended next action} as 0.5 due to the absence of coagulation action recommendation, highlighting a preference for more conservative bleeding control measures.
The second sample revealed broader clinical judgment. A discrepancy emerged between the next surgical action that occurs in the actual video sequence (clipping) and both the AI's and surgeons' preferred next action (dissection). This divergence arose because the operating surgeon adhered to a more lenient interpretation of the Critical View of Safety (CVS) criteria, proceeding to clipping earlier. In contrast, the AI and evaluating surgeons followed stricter CVS standards, deeming further dissection necessary before clipping. From a surgical safety perspective, adhering to stricter safety criteria aligns with the imperative to minimize risk and prioritize patient well-being.

\section*{DISCUSSION}

The development of AI-driven surgical systems has been hindered by the lack of generalizable frameworks capable of interpreting the complex, yet structured, nature of surgical procedures. Our work addresses this challenge by introducing a comprehensive BSA dataset and a foundation model capable of recognizing these basic actions across multiple specialties. 

Our BSA-10 dataset enables robust BSA recognition and automation research, with diverse, high-quality annotations supporting real-time surgical AI development. The universality of BSAs, despite variations in anatomical context, suggests that surgical workflows can indeed be decomposed into reusable, standardized components, enabling more robust and generalizable AI applications in surgery. The BSA foundation model generalizes across surgeons, hospitals, and unseen specialties. Applications include: (i) Skill analysis: Surgical action barcode provides feedback for technique refinement. (ii) Surgical action planning (SAP): MLLM-based SAP models long-horizon BSA chains, enabling procedure automation and intraoperative guidance. Predictions may streamline workflows by handling routine tasks while surgeons focus on critical steps. Textual reasoning can aid education and assessment.

Limitations include restrictions to predefined BSA categories, limiting the discovery of novel surgical actions. Real-time robotic integration remains unexplored. The framework currently lacks continuous learning capabilities essential for adapting to evolving surgical practices. Future directions should prioritize developing adaptive learning mechanisms for organic knowledge base updates and direct robotic system integration for autonomous action execution. The fusion of vision, language, and action recognition promises to revolutionize surgical education, intraoperative guidance, and procedural automation, establishing intelligent surgical ecosystems that enhance precision, safety, and accessibility in minimally invasive procedures.

\section*{MATERIALS AND METHODS} \label{section:methods}
\subsection*{Basic Surgical Action Dataset and Evaluation}
\textbf{Basic surgical action dataset generation pipeline:} 
To create the BSA-10 dataset, we followed a systematic, multi-step approach to ensure its quality and diversity (see Figure~\ref{fig_meta_action_dataset}a): \textbf{Step 1: BSA criteria establishment.} We proposed a standardized BSA annotation protocol for the public and external validation datasets. Beyond fulfilling the description of the BSA, candidate video clips should satisfy (a) Each clip's duration is between 2 and 40 seconds, ensuring that the clips are short enough to focus on individual actions while being long enough to capture meaningful movement and context. Regarding the temporal demarcation of an action, we adopt two criteria: (1) a complete action should ideally include three sub-stages: (i) the instrument's approach to the tissue, (ii) the sustained instrument-tissue interaction, and (iii) the instrument's withdrawal from the tissue; and (2) every annotated action must persist for at least two seconds.
It should be emphasized that the three sub-stages of approach, interaction, and withdrawal cannot always be unambiguously captured. This limitation stems from our primary objective, namely to isolate a video segment that contains exactly one action, and from several practical operations: the instrument may remain in close proximity to the tissue, rendering the approach phase indistinct, or it may sustain prolonged motionless contact with the tissue before another dominant action begins, forcing an early truncation that precludes capture of the withdrawal phase.
(b) The BSA must occur in the center of the surgical field to ensure visibility and context for the action. (c) Each video clip contains only one BSA to maintain clarity and avoid confusion during model training. \textbf{Step 2: Repurpose of existing annotations.} We repurposed existing annotations from publicly available datasets to prepare the BSA annotations. This allowed us to leverage pre-existing work while ensuring the labeling was accurate and consistent. \textbf{Step 3: Annotation of new BSA clips.} We have annotated several publicly available surgical datasets that originally contained raw surgical videos without action labels, requiring us to manually annotate BSA clips directly from the raw videos according to our defined BSA criteria. This process requires significant effort, as it involves meticulously reviewing and annotating each video clip to ensure accurate and consistent labeling of surgical actions. The annotation process is carried out with careful attention to detail, ensuring that each action is identified and labeled according to predefined criteria, thereby enhancing the quality and reliability of the dataset for subsequent analysis and model training. Then we generated the BSA video clips by utilizing the annotation files with manually annotated start and end timestamps for each action. \textbf{Step 4: Overall dataset review.} After collecting the initial clips, we manually checked and cleaned the dataset with surgical expert review to ensure that each clip is accurately labeled and clinically validated. We also supplemented the dataset by manually annotating additional action clips of less frequent actions, such as knot-tying and packaging.

\noindent
\textbf{Protocol for basic surgical actions:}
During the data annotation process, all basic surgical actions are clearly defined with specified descriptions and a concise labeling protocol across surgical videos~\cite{ma2022surgical,nwoye2022rendezvous, cui2024capturing, gao2014jhu,van2021gesture,ahmidi2017dataset, chen2022surgesture}.

\textbf{Aspiration:} removing fluids or tissue debris from the surgical site, typically through suction. Aspiration allows surgeons to ensure a clean environment, facilitating precise tissue manipulation and reducing the risk of contamination or infection.

\textbf{Clipping:} applying surgical clamps to control bleeding or temporarily occlude blood vessels or tissue. The ability to efficiently perform clipping is essential for minimizing blood loss, ensuring both patient safety and the success of the surgery.

\textbf{Coagulation:} applying thermal energy to promote hemostasis by denaturing proteins, forming a stable clot, and sealing tissue. Effective coagulation is crucial for maintaining patient stability throughout the procedure and preventing postoperative complications associated with excessive bleeding.

\textbf{Dissection:} separating or cutting of tissue layers to expose underlying structures during surgery and access deeper anatomical layers. It is fundamental to most surgical procedures as it allows the surgeon to access the targeted area while minimizing damage to surrounding tissues. 

\textbf{Knot-tying:} twisting the suture material in a series of loops and tying a suture to secure tissues together. Tying secure knots with appropriate tension is vital for reducing the risk of wound complications.

\textbf{Needle Grasping:} moving the instrument towards the needle in a smooth, controlled manner and grasping the needle, emphasizing the dynamic motion of ``go to grasp''. Accurate execution of this action allows the surgeon to control the needle and ensure proper tissue alignment precisely. Proper grasping of the needle position is critical to minimizing tissue damage and promoting optimal wound healing.

\textbf{Needle Puncture:} driving a needle to penetrate the tissue and ensuring that the needle enters the tissue at the desired angle and depth. Precision in needle puncture ensures that the needle enters tissues correctly, reducing trauma and improving the efficacy of the repair. 

\textbf{Packaging:} placing tissue or other materials into a sterile bag, minimizing contamination, and maintaining a clean surgical field. This action minimizes contamination risks and helps maintain a clean surgical field by preventing the spread of infectious agents from the surgical site. 

\textbf{Suture Pulling:} drawing the suture material through the tissue after the needle has passed through it. This action is achieved by pulling the needle or thread. Proper suture tension and technique are crucial for minimizing postoperative complications such as wound dehiscence and infection.

\textbf{Tissue Retraction:} holding and manipulating tissues or organs aside to provide better visibility and access to the surgical site. By gently pulling tissues aside, surgeons can expose deeper structures and maintain a clear view during surgery while preventing injury to surrounding organs and tissues.

% H.C. Yip, Y. Liu, R. Mai,  M. Xu, D. Imans, Y. Ye
\noindent
\textbf{Assessment of inter-rater agreement rate:}
The dataset was collaboratively annotated by three experienced raters (Rater Team 0: M. Xu, D. Imans, Y. Ye). For videos with pre-existing action annotations from public datasets, we mapped their original labels to our standardized action protocol. However, these repurposed labels often exhibited coarser temporal granularity. We therefore refined these boundaries through manual review to ensure strict alignment with our predefined action definitions. Videos without prior annotations were manually labeled from scratch using the same action definitions.

The final annotated dataset comprised $8,219$ action clips and was subsequently verified by two specialist physicians: Dr. Y. Liu (urologist, Rater 1) and Dr. R. Mai (hepatopancreatobiliary specialist, Rater 2). We employed Cohen's kappa ($\kappa$)~\cite{hallgren2012computing}, Gwet's AC$_1$ coefficient~\cite{gwet2008computing}, and the Pearson correlation coefficient (PCC)~\cite{benesty2009pearson} to assess inter-rater agreement with $2$ raters. Cohen's kappa measures agreement beyond chance based on raters' marginal distributions, but suffers from known paradoxes when category prevalence is skewed, or agreement is high. Gwet's AC$_1$ addresses these limitations by defining chance agreement according to category prevalence rather than rater-specific tendencies, yielding more stable estimates in such contexts. PCC quantifies the linear association between integer-coded labels, ranging from -1 (perfect negative correlation) to +1 (perfect positive correlation).

Cohen's Kappa ($\kappa$) was calculated according to Equation~\ref{eq:cohen_kappa}, where $P_{o}$ is the observed agreement rate and $P_{e}$ is the expected agreement rate:

\begin{equation}
\kappa = \frac{P_{o} - P_{e}}{1 - P_{e}}
\label{eq:cohen_kappa}
\end{equation}

The Pearson Correlation Coefficient was calculated according to Equation~\ref{eq:pearson}, where $x_i$ and $y_i$ are the individual sample points indexed with $i$, and $\bar{x}$ and $\bar{y}$ are the respective means:

\begin{equation}
r = \frac{\sum_{i=1}^{n}(x_i - \bar{x})(y_i - \bar{y})}{\sqrt{\sum_{i=1}^{n}(x_i - \bar{x})^2}\sqrt{\sum_{i=1}^{n}(y_i - \bar{y})^2}}
\label{eq:pearson}
\end{equation}

Gwet's AC$_1$ coefficient was calculated according to Equation~\ref{eq:gwet_ac1}, where $P_{o}$ is the observed agreement rate, $P_{e}^{\text{AC}_1}$ is the expected agreement rate under independence, and $\pi_j$ is the prevalence of category $j$:

\begin{equation}
\text{AC}_1 = \frac{P_{o} - P_{e}^{\text{AC}_1}}{1 - P_{e}^{\text{AC}_1}}, \quad \text{where} \quad P_{e}^{\text{AC}_1} = \sum_{j=1}^{k}\pi_j(1-\pi_j)
\label{eq:gwet_ac1}
\end{equation}

To evaluate inter-rater consistency, both specialists independently reviewed $8,219$ clips. Overall, $324$ disagreements were observed among the total clips, resulting in an observed agreement of $96.06\%$ and a disagreement rate of $3.94\%$. 
The Pearson correlation coefficient (PCC) between the two raters was 0.8727, indicating strong agreement. The analysis yielded a Cohen's kappa ($\kappa$) score of $0.9435$, consistent with almost perfect agreement~\cite{landis1977measurement}. 
According to Landis and Koch~\cite{landis1977measurement}, Cohen's $\kappa$ coefficients were interpreted using established benchmarks: poor ($\kappa < 0$), slight ($0 \leq \kappa \leq 0.20$), fair ($0.21 \leq \kappa \leq 0.40$), 
moderate ($0.41 \leq \kappa \leq 0.60$), substantial ($0.61 \leq \kappa \leq 0.80$), or 
almost perfect ($\kappa \geq 0.81$). Gwet's AC$_1$ coefficient of 0.9587 indicates excellent inter-rater reliability.

Dr. R. Mai identified 141 clips with disagreed labels, while Dr. Y. Liu identified 247 such clips. The intersection of these disagreements comprised 39 clips, with the union of disagreements calculated as 349 clips. Upon detailed review of the disputed clips, 155 clips were excluded from the dataset. Of the remaining 194 clips requiring modification, 55 underwent label reclassification, whereas 139 were subjected solely to temporal boundary adjustments without categorical alteration. The clean dataset was derived by removing all excluded clips and updating the modified ones from the original corpus, yielding 8,064 clips. 

After resolving the disagreements based on feedback from Rater 1 and Rater 2, Rater 0 (M. Xu) subsequently collected an additional $3851$ action clips, expanding the dataset to $11,915$ clips. These $3851$ supplementary clips were again independently reviewed by Dr. Y. Liu and Dr. R. Mai. Both reviewers, blinded to each other's assessments, evaluated the temporal boundaries and categorical labels of each clip against the pre-defined surgical action protocol. This conservative approach ensured that reviewers were not burdened with re-evaluating previously adjudicated clips, thereby preventing circular validation bias and maintaining the integrity of the independent review process. The observed agreement between annotators was 99.09\%, with only 35 clips (0.91\%) requiring label overrides by Dr. Y. Liu and complete concordance from Dr. R. Mai.
Cohen's kappa ($\kappa$) was calculated to account for chance agreement, yielding a value of 0.9894, which corroborates the excellent agreement between raters. The PCC further supported this strong association, with a value of 0.9746. We additionally report Gwet's AC$_1$ coefficient, which yielded a value of 0.9902 and indicates near-perfect inter-rater reliability. These results demonstrate robust consistency between expert annotators and validate the reliability of the labeling protocol.

Given the substantial dataset size and high costs of manual annotation, and considering that Dr. Y. Liu and Dr. R. Mai had already reviewed the dataset, we randomly sampled $500$ clips from the total video corpus ($1,1915$ clips), ensuring balanced representation across $10$ action classes and $6$ surgery types. Dr. H.C. Yip (a senior urologist, Rater 3) then independently annotated these randomly selected 500 clips. We employed the PCC to assess the inter-observer agreement. The PCC turned out to be 0.9862, which indicates an excellent degree of agreement between the two raters. Additionally, Cohen's kappa was calculated to account for chance agreement, yielding a value of 0.9911, further substantiating the near-perfect concordance between Rater 0 and Rater 3. The observed agreement rate was 99.2\%. Gwet's AC$_1$ coefficient of 0.9912 indicates near-perfect inter-rater reliability.

Consensus decisions, informed by inter-rater feedback, addressed the following scenarios:

(a) For the aspiration class, surgeons clarified that when the suction device is forcibly swept across the tissue surface, scraping rather than simply residing in a pool of fluid, it is performing dissection and should not be labelled as aspiration even if minor blood is visible. The annotator must verify that the primary intent is fluid evacuation and that the tip is visibly aspirating blood or saline.

(b) Because sometimes the same electrocautery tool is used for both coagulation and burn dissection, surgeons adopted a purpose driven rule: if the electrode is applied to achieve haemostasis, that is, the target is a bleeding vessel or vascular pedicle, the segment is labelled coagulation; if the electrode is swept along an anatomical plane to separate tissue, it is labelled dissection regardless of incidental coagulative effect.

(c) We did not attempt to subclassify blunt versus sharp dissection, because the two modes typically alternate in rapid succession and seldom yield discrete, isolatable video segments. Consequently, any sequence of dissection gestures directed at the same anatomical site was annotated as one continuous dissection action.

(d) Within each trimmed video clip, tissue retraction is the only action permitted to co-occur with any of the remaining nine action classes, as tissue retraction is often employed as a supporting maneuver. All other actions must be strictly isolated; simultaneous presence of any two non-retraction categories is disallowed.

(e) The surgical stapler performs integrated tissue apposition and transection (termed ``stapling''), a function distinct from discrete ``clipping''. Based on feedback from the two consulting surgeons, stapling was therefore distinguished from clipping in our annotation schema, and all stapling instances were consequently excluded from the clipping class in the final dataset.

\noindent
\textbf{Long-tail distribution analysis of the dataset:}
Figure~\ref{fig_meta_action_dataset}d illustrates the frequency and distribution of BSAs across various procedures. The analysis shows a long-tail distribution, with dissection occurring most frequently, while knot-tying and packaging are among the least observed actions in the dataset. Indeed, during the surgery, most instruments often dissect tissues, which explains the higher frequency of this action. The knot-tying and packaging actions present lower frequencies as they only happen in very particular moments of the surgery. While needle puncture and suture pulling may appear to be symmetrical actions at first glance, suture pulling is less frequent than needle puncture for the following reasons: (a) In certain surgical areas, the time gap between one ``needle puncture'' and the next puncture is very short. In such cases, the ``suture pulling'' action may not be a distinct movement, but rather a brief transition as the surgeon shifts to perform the subsequent puncture. (b) During suturing, the length of the suture decreases as the procedure progresses, resulting in a shorter duration for each action and making it more challenging to capture effective clips of the ``suture pulling'' action.

\noindent
\textbf{Classification metrics for the BSA recognition model:}
For each action class $k$, we evaluate the performance of the model across multiple decision thresholds, systematically varying the threshold applied to the predicted confidence scores. At each threshold, we compute the following classification metrics:
\begin{itemize}
    % \item \textbf{True Positives} (${TP_{k}}$): Correctly identified instances of action $k$.
    \item \textbf{True Positives} (${TP_{k}}$): The number of instances of action $k$ correctly identified by the model.
    \item \textbf{False Positives} (${FP_{k}}$): Instances incorrectly predicted as action $k$, but belonging to a different class.
    \item \textbf{True Negatives} (${TN_{k}}$): Instances correctly identified as not belonging to action $k$.
    \item \textbf{False Negatives} (${FN_{k}}$): Instances of action $k$ that were incorrectly classified as a different class.
\end{itemize}

Based on the computed values at each threshold, we derive the sensitivity and specificity as:

\begin{equation}
    \text{Sensitivity}_{k} = \frac{{TP_{k}}}{{TP_{k}}+{FN_{k}}}
\end{equation}

\begin{equation}
    \text{Specificity}_{k} = \frac{{TN_{k}}}{{TN_{k}}+{FP_{k}}}
\end{equation}
These values are then plotted to construct the ROC curve.
The optimal threshold is selected as the point that maximizes the Youden Index $J$, which is computed as:
\begin{equation}
    J = \max \left( \text{Sensitivity} + \text{Specificity} - 1 \right)
\end{equation}
This threshold balances both sensitivity and specificity, ensuring that neither false positives nor false negatives disproportionately affect the action classification. Mathematically, the selected threshold $\tau_k$ is given by:
\begin{equation}
    \tau_k = \arg\max_{\tau} \left( \text{Sensitivity}_{k}(\tau) + \text{Specificity}_{k}(\tau) - 1 \right)
\end{equation}

This process is repeated for each action class $k$. Then each sample will be recomputed according to the new class-specific thresholds, yielding $\hat{TP_{k}}$, $\hat{TN_{k}}$, $\hat{FP_{k}}$ and $\hat{FN_{k}}$, which are then used to calculate the corrected specificity $\frac{\hat{TN_{k}}}{\hat{TN_{k}}+\hat{FP_{k}}}$ and sensitivity $\frac{\hat{TP_{k}}}{\hat{TP_{k}}+\hat{FN_{k}}}$ of each action class to keep consistent with the Youden Index and ROC curve.

\subsection*{General-purpose Model for Basic Surgical Action Recognition across Procedures}
In this study, we develop a BSA foundation model, a Transformer-based deep learning model with spatiotemporal reasoning capability for the automated recognition of basic surgical actions from endoscopy videos.

Given an input video clip $ \mathnormal{x} \in \mathbb{R}^{H \times W \times 3 \times T} $, which consists of $ T $ RGB frames of spatial dimensions $H \times W$, we define the action recognition model as a function $\mathcal{R}$ that maps the clip to a probability distribution over a predefined set of ten basic surgical actions $\mathcal{A}$. The predicted action label $a$ is obtained by selecting the class with the highest predicted probability:

\begin{equation}
a = \underset{i \in \mathcal{A}}{\arg\max} \, \mathcal{R}(x)_i,
\end{equation}
where the action set $\mathcal{A}$ = \{Aspiration, Coagulation, Dissection, Knot-tying, Needle Grasping, Needle Puncture, Packaging, Suture Pulling, Tissue Retraction, Vessel Clipping\}, and $\mathcal{R}(x)_i$ denotes the predicted probability of clip $x$ belonging to action $i \in \mathcal{A}$.

\begin{figure*}[t!]
\centering
\includegraphics[width=1\linewidth]{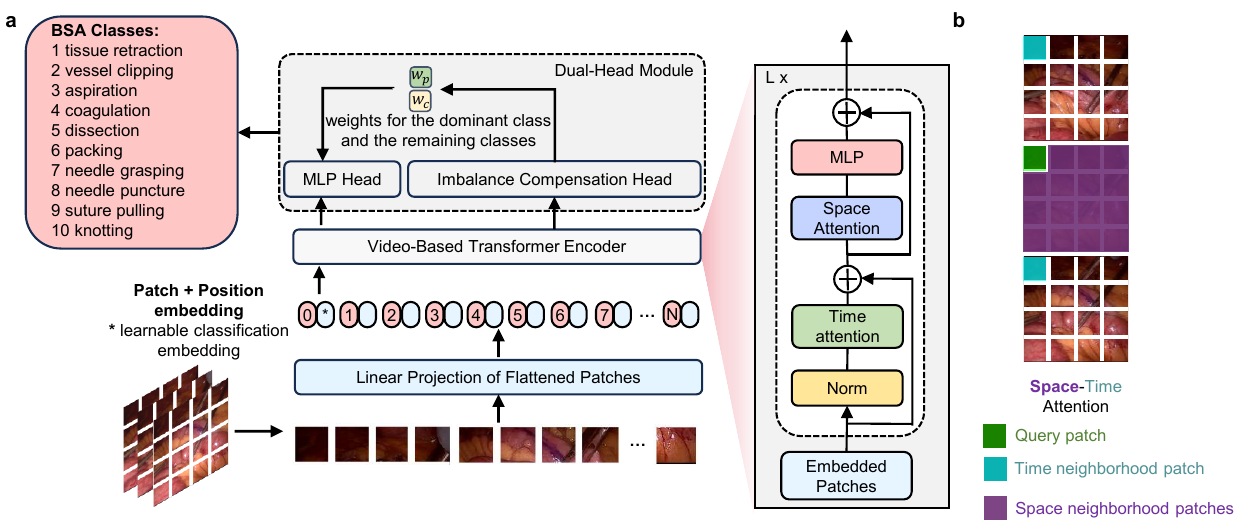}
\caption{{\bf BSA foundation model.} \textbf{a}
Each frame from the action clip is divided into patches, which are subsequently transformed through linear embedding. These flattened patches are then fed into the video transformer encoder, followed by processing through a dual-head architecture. The dual head comprises an MLP head and a newly proposed imbalance compensation head. The latter is designed to refine class predictions by applying weights to the output probabilities generated by the MLP, thereby addressing disparities in class distributions. The video self-attention blocks involve sequentially applied temporal and spatial attention layers that perform self-attention over a defined spatiotemporal region of frame-level patches. We employ residual connections to integrate information across different attention layers within each block. At the end of each block, a single-hidden-layer MLP is applied. The complete model is built by stacking these blocks sequentially.
\textbf{b}
Visualization of the space-time self-attention mechanism applied to a video example. 
}
\label{fig_classification_model}
\end{figure*}

The architecture of our BSA foundation model is depicted in Figure~\ref{fig_classification_model}a. Following the approach of ViT~\cite{dosovitskiy2021imageworth16x16words}, each video frame is divided into $ N $ non-overlapping patches of size $P \times P $, where $ N = \frac{HW}{P^2}$. 
We use a learnable matrix $M \in \mathbb{R}^{D \times (P^2 \cdot 3)}$ to project each flattened patch (with $3$ representing RGB channels) into a $D$-dimensional latent space. To encode the spatiotemporal position of each patch, we add a learnable positional embedding $e_{(p,t)}^{pos}$, where 
$p$ and $t$ index the spatial patch and temporal frame, respectively. Each embedded patch is thus represented as:
\begin{equation}
\begin{aligned}
z_{(p,t)} = M x_{(p,t)}+e_{(p,t)}^{pos}. 
\end{aligned}
\end{equation}

The resulting sequence of patch embeddings $z_{(p, t)}$ is prepended with a special learnable classification token $z_{(0, 0)}$, which aggregates global information during transformer processing and serves as the class token. This sequence is then passed to a video-based Transformer encoder, functioning analogously to token embeddings in language models for natural language processing (NLP) tasks.

\noindent
\textbf{Video-based transformer encoder with space-time attention:}
The video-based transformer encoder (Figure~\ref{fig_classification_model}b) aims to extract rich feature representations from the patch embeddings $z$. 
In the encoder, these embeddings are first normalized to facilitate stable training and are then processed through sequential temporal and spatial attention mechanisms to effectively model spatiotemporal dependencies, such as instrument movements and tissue interactions. 
The features are progressively refined through L repeated encoding blocks. The space-time self-attention mechanism follows the standard self-attention formulation~\cite{dosovitskiy2020image}, using queries (Q), keys (K), and values (V) to capture relationships across time and space. Temporal attention is calculated for each patch at position $(p,t)$ by attending to all patches located at the same spatial position $p$ across different frames. The output of temporal attention is then passed to the spatial attention module, which computes attention among all patches within the same frame $t$. 
This process performs only $N+T+2$ comparisons per patch, where $N$ denotes the number of spatial locations, $T$ denotes the number of frames, and the additional $+2$ accounts for the class token and the learnable relative positional bias, resulting in an overall computing complexity of $\mathcal{O}((N+T+2)d^2)$ compared to $\mathcal{O}(NTd^2)$ for full spatiotemporal attention, where $d$ is the feature dimension.

Figure~\ref{fig_classification_model}a illustrates the overall architecture of our spatiotemporal attention, and Figure~\ref{fig_classification_model}b provides how temporal and spatial attention mechanisms are applied to a video example. Each video clip is represented as a sequence of frame-level patches, where each patch corresponds to a $16 \times 16$ pixel region. In the visualization, the query patch is highlighted in green, and its corresponding space-time attention neighborhood is depicted in non-green colors. Patches that remain uncolored are excluded from the self-attention computation~\cite{vaswani2017attention} for the query patch. Purple and blue colors indicate separate attention operations applied across spatial and temporal dimensions, respectively. 
Notably, self-attention is computed independently for every patch in the video clip, meaning each patch serves as a query at some point during the process. Although Figure~\ref{fig_classification_model}b illustrates attention patterns for only two consecutive frames, the same attention scheme is applied across all frames in the clip.

\noindent
\textbf{Dual-head module for class balancing with penalization and compensation weights:}
The presence of a dominant class in the dataset often leads to challenges in model prediction. Basic surgical actions that are less distinct or harder for the model to recognize tend to increase prediction uncertainty, causing the model to favor the dominant class due to its higher estimated probability. Although this tendency may improve overall accuracy, it compromises the recognition on minority classes. 

To address this imbalance issue, a more refined prediction strategy is essential for achieving balanced and fair performance across all classes. To this end, we propose a dual-head architecture that combines a standard multi-class MLP head with an imbalance compensation head, designed to mitigate biases arising from uneven class distributions. 
The imbalance compensation head specifically targets the dominant class, the most prevalent category in the dataset. This head generates a penalization weight $ w_{\text{p}} $ and a compensation weight $ w_{\text{c}} $ for the dominant class and the other classes. These weights are computed from the input feature representation via a fully connected layer:
\begin{equation}
\begin{aligned}
% \[
w_{\text{p}},w_{\text{c}} = f_{\text{imbalance}}(\theta),
% \]
\end{aligned}
\end{equation}
where $ f_{\text{imbalance}}(\theta) $ denotes the imbalance compensation head.

The original multi-class prediction probabilities from the MLP head are denoted as: $\bm{\phi}_{\text{MLP}} = [\phi_{\text{dominant}}, \phi_2, \dots, \phi_c]$, where $ c $ is the total number of classes, $\phi_{\text{dominant}}$ is the predicted probability for the dominant class, and $ \phi_i $ is the predicted probability for class $ i $, where $i > 2$. 
To mitigate the class imbalance, the final adjusted probabilities from our dual-head module $ \bm{\phi}_{\text{dual-head}} $ are weighted by the penalization weight $w_{\text{p}}$ and compensation weight $w_{\text{c}}$ as follows:

\begin{equation}
\begin{aligned}
\bm{\phi}_{\text{dual-head}} = [w_{\text{p}} \cdot \phi_{\text{dominant}}, w_{\text{c}} \cdot \phi_2, \dots, w_{\text{c}} \cdot \phi_c].
\end{aligned}
\end{equation}
This reweighting mechanism helps suppress the influence of the dominant class and compensates for the underrepresented ones, to encourage more balanced predictions across all basic surgical action categories.

\noindent
\textbf{Loss function:}
To further mitigate the influence of the dominant class in uncertain cases and enhance prediction reliability, we employ the Evidential loss~\cite{sensoy2018evidential} during training. This loss function encourages well-calibrated predictions by modeling uncertainty through a Dirichlet distribution framework, avoiding both overconfidence and excessive uncertainty.

\noindent
\textbf{Implementation details and hyperparameters:}
Based on the manually annotated start and end timestamps for each action, we create the corresponding action video clip. For each action video clip, we first downsample it to 1 fps. Before feeding the clip into the neural network, we sample it with a frame interval of 4, uniformly retaining 16 frames. For example, for an action originally lasting 33 frames, we first loop it to form a 64-frame sequence [1, 2, ..., 33, 1, 2, ..., 31], and then sample every 4 frames, resulting in the final frame sequence [1, 5, ..., 33, 4, 8, ..., 28]. For a video originally longer than 64 frames, during training, we randomly select the starting point to ensure sample diversity; during inference, we use the central frame of the original video as the starting point to ensure inference stability. Multiple experiments have shown that this setup helps maintain the balance between computational complexity and the preservation of visual information in the video.

We implement all experiments in Pytorch with four NVIDIA RTX 4090 GPUs. Each RGB frame is resized to 224$\times$224. The ViT-based feature extraction backbone employs the pretrained weights of ViT-B/16 and generates an embedding of dimension \textit{D}=768 for each video clip. The embedding is then fed into both the classification head and the imbalance compensation head to produce the final classification probability. The model is optimized using stochastic gradient descent (SGD) with the initial learning rate of 0.005, momentum of 0.9, weight decay of 0.001, and a batch size of 6. Training is performed for 50 epochs and takes around 5 hours to complete.

\noindent
\textbf{Ten-fold cross-validation and data splits:}
We evaluate the developed BSA foundation model using ten-fold cross-validation, where each fold's test set consisted of action clips from videos unseen during training. Specifically, we perform a video-level random split, resulting in ten folds with the following numbers of action clips: 1,067, 1,439, 1,066, 1,451, 1,073, 1,284, 1,334, 1,069, 1,093, and 1,039 clips, respectively.
This video-level splitting method (as opposed to random clip-level splitting) creates a more challenging evaluation scenario because it prevents information leakage from correlated clips of the same video and better tests generalization to new surgical scenes from different patients. This video-level splitting method supports the evaluation of the BSA recognition model's ability to generalize to previously unseen videos, more closely reflecting real-world application scenarios. 

The precise distribution of individual BSA classes across all ten cross-validation folds is presented in the supplementary materials Table~\ref{tab:fold_act}. The data reveal the inherent class imbalance within surgical procedures, with dissection representing the most frequent action (5,341 video clips) due to its fundamental role across multiple surgical phases, while specialized actions like packaging (148 video clips) and knot-tying (504 video clips) occur less frequently as they correspond to specific procedural moments. This distribution reflects authentic surgical workflow patterns where certain basic actions naturally dominate due to their essential and repetitive nature during minimally invasive procedures.

\subsection*{BSA for Surgical Action Planning with Multimodal Large Language Model} 
By decomposing complex surgical procedures into basic surgical actions, we establish a multi-modal vision-language-based action planning framework to predict the next surgical action and connect these actions into long-horizon chains, leveraging the advanced multimodal capabilities of GPT-4o, a state-of-the-art Multimodal Large Language Model (MLLM). The MLLM is enhanced with expert-provided surgical domain knowledge integrated as system prompts, including surgical process, safety protocol, and action description, as illustrated in the upper-left portion of Figure~\ref{planning_results}.
The MLLM generates a historical analysis by processing distant history actions predicted by the BSA foundation model, the recent history action clip with its associated prediction from the BSA foundation model, and the current observation represented by the single frame. This reasoning output covers multiple dimensions, such as scene understanding, progress judgment, and safety considerations. Utilizing this analysis, the model subsequently recommends the upcoming basic surgical action.
This capability, enabled by novel prompt engineering techniques, ensures context-aware predictions of future basic surgical actions while adhering to safety and procedural guidelines, as illustrated in Figure~\ref{planning_results}.

\noindent
\textbf{Task definition:}
For the Surgical Action Planning (SAP) task, two key inputs are provided to facilitate the generation of a future action plan:

\textbf{Visual history $C$}: A video that offers contextual information on the progress made toward a goal from the beginning up to the current time point, denoted as $t$. $C$ encompasses $n$ actions that contribute to achieving the goal. 

\textbf{Goal prompt $G$}: A series of prompts using the natural language description designed to achieve the goal. Examples include: ``You are a surgical assistant specializing in laparoscopic cholecystectomy, with expertise in the following: surgical workflow, safety protocols, and procedural action descriptions. What is the next action to advance the dissection of Calot's Triangle during a laparoscopic cholecystectomy procedure?''.

Surgical action planning (SAP) task $\varphi$ can be conceptualized as a forward-looking sequential decision-making problem, where the model predicts future actions by following a policy that is conditioned on the historical context $C$ (which acts as a partially observed state) and the goal prompt $G$.

\begin{equation}
\begin{aligned}
    \varphi  = \mathcal{F}(a_{t+1}, a_{t+2}, ...|C,G)
\end{aligned}
\end{equation}

\noindent
\textbf{Metrics description:}
Building upon our definitions of local and global accuracy presented in the Results section, we further explain these metrics in detail.

\textbf{\textit{Strict Local Accuracy (S-LocalAcc):}} A model's predicted action $\hat{a}_i$ is considered correct only if it exactly matches the ground truth $a_i$ in both action class and temporal location. We also consider Top-2 and Top-3 accuracy to allow for more flexible evaluation, where a prediction is accepted if the true action appears within the model's two or three highest-confidence predictions, respectively.  

\textbf{\textit{Strict Global Accuracy (S-GlobalAcc):}} To assess generalization across complete surgical contexts, we compute \textit{S-GlobalAcc} (video-level accuracy) through hierarchical evaluation of strict binary correctness:
\begin{equation}
\textit{S-GlobalAcc} = \frac{1}{M} \sum_{i=1}^{M} \left( \frac{1}{Q_i} \sum_{j=1}^{Q_i} \mathbb{I}(\hat{a}_{ij} = a_{ij}) \right)
\end{equation}

where $M$ is the total number of patient contexts, $Q_i$ is the number of samples in $i^{th}$ patient context, $\hat{a}_{ij}$ represents the predicted action for $j^{th}$ sample in $i^{th}$ context, $a_{ij}$ represents the ground truth action, $\mathbb{I}(\cdot)$ is the indicator function (1 if correct, 0 otherwise).

This provides a macro-level performance measure, ensuring the model's consistency throughout a complete clinical workflow rather than on isolated samples.

\textbf{\textit{Relaxed Local Accuracy (R-LocalAcc):}} Recognizing the inherent temporal variability in surgical workflows, we introduce a relaxed accuracy metric. Here, a prediction $\hat{a}_i$ is considered correct if it matches either the current ground truth action $a_i$ or the subsequent action $a_{i+1}$ (i.e., $\hat{a}_i \in \{a_i, a_{i+1}\}$). This +1-step tolerance accounts for clinically permissible variations in action sequencing without compromising procedural integrity.  

\textbf{\textit{Relaxed Global Accuracy (R-GlobalAcc):}} By computing accuracy within each entire patient context under the +1-step tolerance ($\frac{1}{Q_i}\sum_{j=1}^{Q_i}$) and averaging these results across all patient contexts ($\frac{1}{M}\sum_{i=1}^{M}$), we obtain the patient-level global accuracy. 

\begin{equation}
\textit{R-GlobalAcc} = \frac{1}{M} \sum_{i=1}^{M} \left( \frac{1}{Q_i} \sum_{j=1}^{Q_i} \mathbb{I}(\hat{a}_{ij} \in \{a_{ij}, a_{(i+1)j}\}) \right)
\end{equation}

\noindent
\textbf{Our MLLM-based visual planning assistance:}
We decompose the planning approach $\varphi$ into two modules. The first module is \textit{BSA recognition module} represented by a function $\mathcal{R}$, which converts a sequence of video history clips $C_{t} = \left \{ x_{1},..., x_{t} \right \} $ into the BSAs that occurred in the video clips $A_{t} = \left \{ a_{1},..., a_{t} \right \}$. The second module enables \textit{forecasting} indicated by a function $\mathcal{F}$, i.e., transforming the output of the recognition module and generating the action plan. A probabilistic formulation of this decomposition can be expressed as:

\begin{equation}
\begin{aligned}
\varphi =\sum_{S_{t}}^{ } \underbrace{\mathcal{F}(a_{t+1}, a_{t+2},...|C_{t}\circ A_{t}, G)}_{forecasting}\underbrace{\mathcal{R}(A_{t}|C_{t})}_{recognition} 
\end{aligned}
\label{equ_overall}
\end{equation}

Using Eq.~\ref{equ_overall} as the foundational expression, we define the input and output for both modules. Subsequently, we present the technical details of the two modules.

\noindent
\textbf{Video action recognition module:}
For the history video clips $C_{t}=\left \{ x_{1}, x_{2},...,x_{t} \right\}$, we use our BSA foundation model to recognize the most probable action for each video clip,  which can be denoted as $A_{t}=\left \{a_{1}, a_{2},..., a_{t} \right \}$. The resultant history segment is termed $S_{t}=\left ( \left ( x_{1},a_{1} \right ), ...,\left ( x_{t},a_{t} \right ) \right )$, where each segment consists of a clip and action.

\noindent
\textbf{Action forecasting module using prompt engineering:}

The future action forecasting module leverages GPT-4o, the multimodal large language model (MLLM), to generate action planning for upcoming procedures, utilizing our innovative historical data development and advanced prompt engineering techniques.
It is worth noting that commercial, closed-source models are common in the research community. These models deliver zero-shot performance in specific cases and facilitate task-specific capabilities through prompt engineering, thereby broadening their potential for specialized applications. Previous studies in MLLM~\cite{liang2023code, vemprala2024chatgpt, brohan2023can, yoneda2024statler,ren2023robots, dai2024think} have predominantly utilized the GPT series models. 
Next, we provide a detailed description of how to develop the framework by utilizing our innovative historical data development and advanced prompt engineering techniques.

\textbf{Development of the history memory module $C$}:
Previous MLLM-based planner typically relies on a set of historical actions and their corresponding clips to comprehend the context. This process is formally expressed as follows:
\begin{equation}
\begin{aligned}
\mathcal{F}(a_{t+1}, a_{t+2},...|\left ( \left ( x_{1},a_{1} \right ), ...,\left ( x_{t},a_{t} \right ) \right ), G)
\end{aligned}
\end{equation}

Prior research has indicated that the approach, which involves extensive analysis of all historical context, can overwhelm the planner, leading to difficulties in detecting state transitions and potential planning errors. The inclusion of too much historical data can introduce unnecessary complexity, diminishing both the efficiency and clarity of the decision-making process. 
In planning tasks, the appropriate next action is influenced by the progression of the surgical procedure and the very recent state of the target tissue.
To address these issues, we propose a refined and novel method for historical contextual understanding that reduces the emphasis on distant historical context while prioritizing more adjacent and relevant details. Specifically, we suggest that the distant history be limited to actions from prior surgical procedures, which can be denoted as $\left (a_{t-4},  ..., a_{t-1} \right )$ predicted by our BSA foundation model, while the near history focuses on the clip-action pair corresponding to the most recent action, which can be represented as $\left (x_{t}, a_{t} \right )$. To further emphasize the most recent scene, the current observation, represented by a single frame $f$, is also included, as shown in Figure~\ref{planning_results}. This revised approach can be formalized as:

\begin{equation}
\begin{aligned}
\mathcal{F}(a_{t+1}, a_{t+2},...|\underbrace{ a_{t-4}, ..., a_{t-1}}_{distant}, \underbrace{ \left (x_{t},a_{t} \right ) }_{near}, f, G)
\end{aligned}
\end{equation}

The proposed historical context development method aims to focus the model's attention on the most relevant, up-to-date information while simplifying the representation of distant historical events, which can reduce the cognitive burden on the model.

To maximize data utilization, we generate multiple input samples from each context segment using a sliding window approach, where each sample shifts forward by one video clip at a time. The sliding window size is $L$. Each sample consists of $L$ historical video clips followed by one current frame. With the shift step of the $1$ video clip, each new video clip is added to the historical context while the oldest video clip is dropped. This ensures that the sample is constantly evolving and contains the most recent history (i.e., the most recent window of $L$ historical video clips) for planning.

\textbf{Development of the goal prompts $G$}:
Along with the development of the historical context, we also provide a surgical domain knowledge base as the system prompts $G_{s}$ for MLLM, which encompasses the \textit{surgical process}, \textit{safety protocol}, and \textit{action description}, as shown in Figure~\ref{planning_results}. \textit{Surgical process} defines the standardized sequence of steps and procedures performed during an operation. \textit{Safety protocol} establishes rules and measures to minimize risks during medical interventions. \textit{Action description} explains individual actions in detail.
User prompts $G_{u}$ guide the system's responses with specific queries, including \textit{scene understanding}, \textit{progress judgment}, and \textit{safety considerations}. Specifically, \textit{Scene understanding} analyzes visual and contextual elements, such as key structures and spatial relationships, to establish the current state. \textit{Progress judgment} dynamically tracks task advancement against the planned workflow, enabling real-time adjustments. \textit{Safety considerations} continuously evaluate risks to prioritize harm-minimizing actions.

These user prompts include the following examples:
(1) \textit{``Please summarize the content of Scene Understanding.''}
(2) \textit{``Please summarize the progress judgment.''}
(3) \textit{``What's safety considerations?''}
(4) \textit{``What should be the next action?''}
(5) \textit{``Please tell me the other two options of actions.''}
The approach is formally expressed as follows:
\begin{equation}
\begin{aligned}
\mathcal{F}(a_{t+1}, a_{t+2},...|\underbrace{ a_{t-4}, ..., a_{t-1}}_{distant}, \underbrace{ \left (x_{t},a_{t} \right ) }_{near}, f, \left ( G_{s}, G_{u} \right ) )
\end{aligned}
\end{equation}

The output is generated as a direct response to these user prompts, delivering a structured and context-aware analysis tailored to the current surgical scenario, followed by a recommendation of the next action that supports task execution and safety, along with reasons for the recommendation.

% \noindent

% \bibliography{iclr2021_conference}
% \bibliographystyle{iclr2021_conference}

% \clearpage

% \section*{SUPPLEMENTARY MATERIALS}

\noindent

\bibliography{iclr2021_conference}
\bibliographystyle{iclr2021_conference}

\clearpage

\section*{ACKNOWLEDGEMENTS}
We would like to express our sincere gratitude to Fengyue Guo (Department of Computer Science and Engineering, The Chinese University of Hong Kong) and An Wang (Department of Electrical and Electronic Engineering, The Chinese University of Hong Kong) for their invaluable assistance in dataset annotation and insightful discussions

\noindent
\textbf{Funding:}
This research work was supported in part by the Research Grants Council of the Hong Kong Special Administrative Region, China, under Projects No. 24209223, No. N\_CUHK410/23, No. T45-401/22-N, and in part by the National Natural Science Foundation of China under Project No. 62322318, and in part by the Ministry of Education Tier 2 grant, NUS, Singapore, under Project No. T2EP20224-0028.

\noindent
\textbf{Conflict of interest}
The authors declare no competing interests.

\noindent
\textbf{Ethics approval and consent to participate}
Ethical approval was obtained from the National Healthcare Group (NHG), Domain Specific Review Board (DSRB) with reference number of 2024/00007.

\noindent
\textbf{Consent for publication}
All authors have reviewed the final manuscript and consent to its publication.

% \noindent
% \textbf{Data availability}
% All the materials and source codes that can reproduce the results in this paper will be open-sourced at \url{https://github.com/XuMengyaAmy/BSA}.

% \noindent
% \textbf{Materials availability}
% All the materials that support the findings of this work are presented in the paper.

% \noindent
% \textbf{Code availability} 
% All the materials and source codes that can reproduce the results in this paper will be open-sourced at \url{https://github.com/XuMengyaAmy/BSA}. 

\noindent
\textbf{Author contribution}
Q. Dou and M. Xu conceived the study. 
Q. Dou, Y. Jin, M. Xu, D. Shen, and J. Zhang designed the work. 
M. Xu, D. Shen, and J. Zhang developed the methodology and conducted experiments.
Q. Dou, Y. Jin,  M. Xu, D. Shen, J. Zhang, H.C. Yip, Y. Gao, C. Chen, D. Imans, H. Ren, Y. Ban,  G. Wang, F. Wong, C.F. Ng, K.Y. Ngiam, R.H. Taylor, and D. Xu analyzed the data.
M. Xu, Q. Dou, H.C. Yip, Y. Gao, G. Wang, F. Wong, C.F. Ng, K.Y. Ngiam, and R.H. Taylor proposed ten basic surgical actions.
M. Xu, D. Imans, Y. Ye, H.C. Yip, Y. Liu, R. Mai, Y. Gao, G. Wang, F. Wong, C.F. Ng, K.Y. Ngiam, and R.H. Taylor constructed the dataset.
H.C. Yip, Y. Gao, G. Wang, F. Wong, C.F. Ng, and K.Y. Ngiam provided clinical perspectives.
Y. Long, H.C. Yip, Y. Gao, K. Chen, G. Wang, F. Wong, C.F. Ng, and K.Y. Ngiam selected the surgical context to implement the action planning.
M. Xu, Q. Dou, Y. Jin, C. Chen, and D. Shen co-wrote the initial manuscript, with all co-authors providing constructive comments and editing. J. Zhang, D. Imans, and Y. Ye conducted this work when they were visiting researchers at The Chinese University of Hong Kong.

%%%%%%%%%%%%%%%% END OF MAIN TEXT %%%%%%%%%%%%%%%

%%%%%%%%%%%%%%%% START OF SUPPLEMENT %%%%%%%%%%%%%%%
\newpage
% 重定义表格和图的编号
\renewcommand{\thetable}{S\arabic{table}}
\renewcommand{\thefigure}{S\arabic{figure}}

% 重置计数器，确保编号从 S1 开始，而不是接着主文的编号
\setcounter{table}{0}
\setcounter{figure}{0}

\section*{Supplementary Materials}
This supplementary material presents extended data in the following key areas:

(1) Table~\ref{tab:surgery_public_dataset} summarizes the six surgery types and their corresponding public datasets. Table~\ref{tab:action_surgery} shows the distribution of surgical actions by procedure type, confirming that the 10 curated action classes are consistently applied across these surgical categories. The ten-fold cross-validation splits are detailed in Table~\ref{tab:fold_act}, which shows the distribution of BSA action classes, and Table~\ref{tab:fold_surg}, which provides the per-fold sample distribution across surgery types.

% The ultimate dataset consisted of 11,915 surgical video clips distributed across ten-fold cross-validation splits, with per-fold sample sizes ranging from 1,073 (fold5) to 1,439 (fold2) (Table~\ref{tab:fold_act}). 

(2) Multi-national external validation demonstrates the model's performance across unseen surgical types from China and Singapore, with Youden Index (95\% CI) quantification for ten basic action classes in both lobectomy and hepatectomy procedures, showcasing cross-regional generalizability; Table~\ref{tab_supp_1} presents detailed performance metrics for lobectomy procedures across all ten BSA classes. The model demonstrates good performance with AUROC values ranging from 79.11\% to 93.77\%. Packaging achieves the highest performance (AUROC: 93.77\%, sensitivity: 86.67\%, specificity: 93.96\%), consistent with main dataset findings due to its distinctive visual characteristics. Coagulation presents the most challenging recognition scenario (AUROC: 79.11\%, sensitivity: 80.00\%), reflecting the complexity of instrument manipulation in confined thoracic spaces. 

Table~\ref{tab_supp_2} demonstrates the model's adaptability to hepatectomy procedures from Singapore, achieving AUROC values ranging from 80.03\% to 97.07\%. While overall performance is lower than lobectomy results, this reflects the combined challenges of both novel procedural context and cross-national validation. Packaging achieves the strongest performance (AUROC: 97.07\%), while needle grasping and tissue retraction show the challenging recognition (AUROC: 79.58\% and 80.03\%). For clipping, our model encountered hepatic-specific technical variations, including the Pringle maneuver requiring temporary vascular clamps rather than permanent clips. Similarly, hepatectomy dissection involves specialized "crush-clamp" techniques, creating distinct visual patterns compared to electrocautery-based separation in other procedures. Our model achieved AUROCs of 97.05\% for clipping and 95.07\% for coagulation. These results demonstrate that while organ-specific requirements and regional surgical preferences introduce visual complexities, BSA fundamental characteristics remain recognizable across international surgical practices.

\begin{figure*}
\centering
\includegraphics[width=0.8\linewidth]{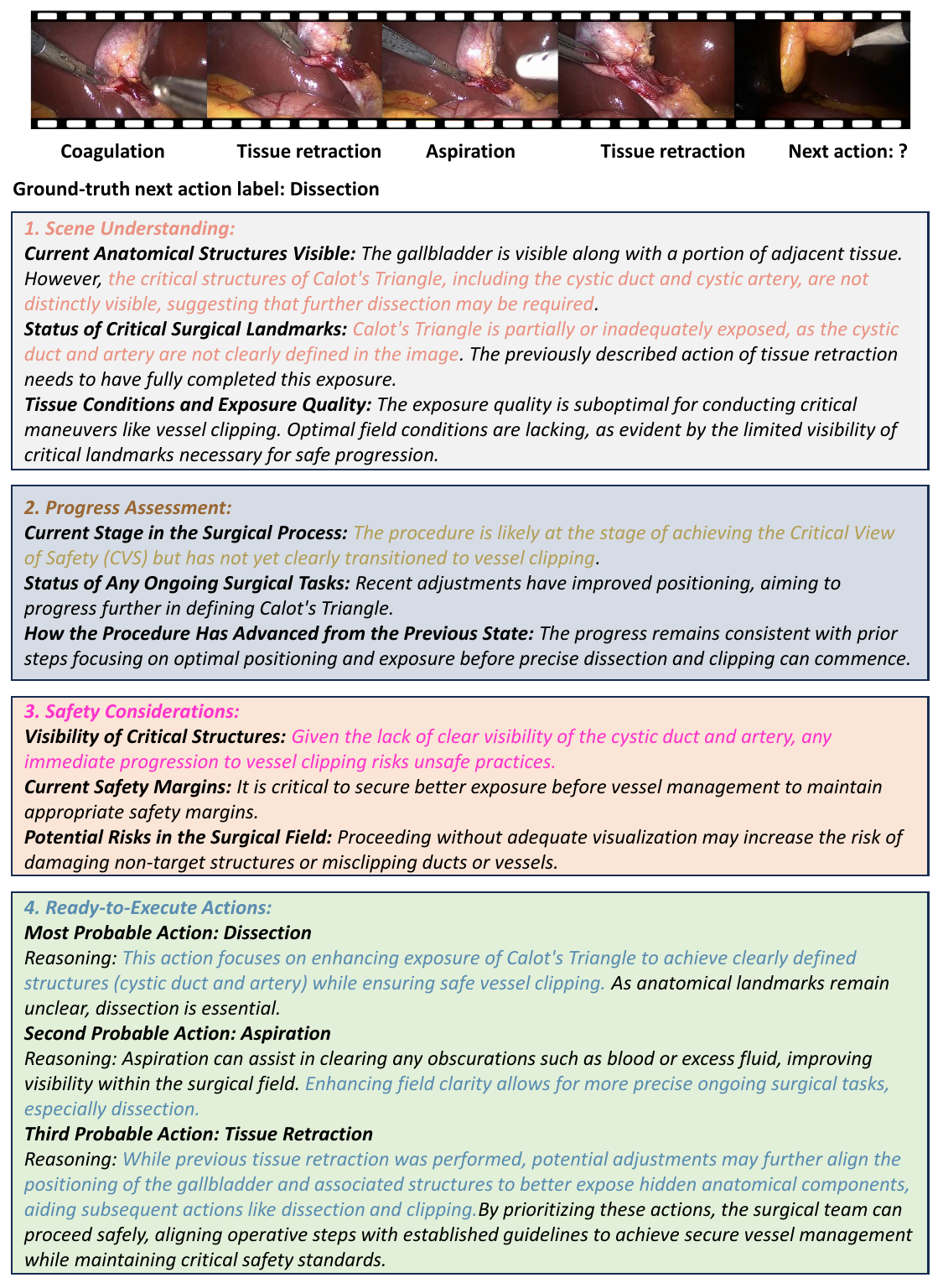}
\caption{{\bf AI Agent's comprehensive surgical reasoning example 1 from C-CVS scenario.}}
\label{fig_supp_1}
\end{figure*}

\begin{figure*}
\centering
\includegraphics[width=0.8\linewidth]{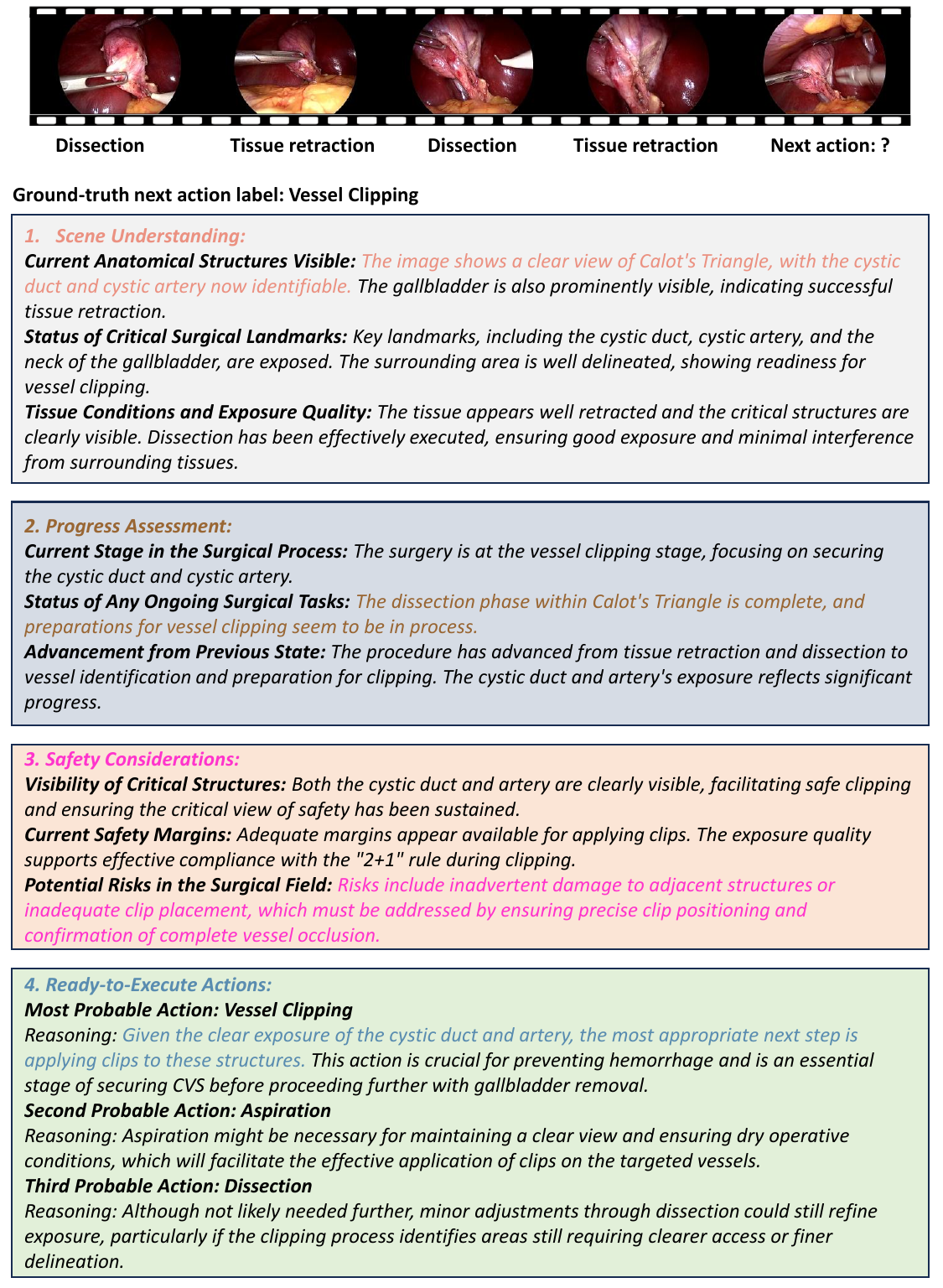}
\caption{{\bf AI Agent's comprehensive surgical reasoning example 2 from C-CVS scenario.}}
\label{fig_supp_4}
\end{figure*}

(3) Exemplars of our AI agent's reasoning process through multimodal large language model-based surgical procedure planning, with detailed textual explanations. Figure~\ref{fig_supp_1}-Figure~\ref{fig_supp_4} showcase comprehensive examples of our AI agent's surgical reasoning capabilities. These detailed analyses demonstrate the model's ability to integrate visual context, procedural knowledge, and safety protocols to generate clinically relevant surgical recommendations. Each example illustrates the complete reasoning process, including scene understanding, progress assessment, safety considerations, and next action recommendations with detailed clinical rationale.

% \textit{Datasets ending with ``yt'' indicate videos collected from YouTube (links in Note)}.

\begin{table*}[ht]
\centering
\caption{Surgery types and the corresponding public datasets and YouTube video sources (SurgYT).}
\label{tab:surgery_public_dataset}
\resizebox{\linewidth}{!}{
\begin{tabular}{@{}ll@{}}
\toprule
Surgery type & Dataset Name \\
\midrule
Cholecystectomy & CholecT50~\cite{nwoye2022rendezvous}, Cholec80~\cite{twinanda2016endonet}, M2cai16-workflow~\cite{cadene2016m2cai}, Heichole~\cite{wagner2023comparative}, SurgYT,   \\
Nephrectomy & Endovis17~\cite{allan20202018}, Endovis18~\cite{allan20202018}, SurgT~\cite{cartucho2024surgt}, Nephrec9~\cite{penza2018endoabs},  SurgYT,  \\
Gastrectomy & MultiBypass140~\cite{ramesh2023weakly}, SurgYT \\
Hysterectomy & Surgicalactions160~\cite{schoeffmann2018video}, Autolaparo~\cite{wang2022autolaparo}, SurgYT,  \\
Prostatectomy & MesadReal~\cite{bawa2021saras}, SAR-RARP50 2022~\cite{psychogyios2023sar}, GraSP~\cite{ayobi2024pixel}, SurgYT \\
Intestinal Resection & HeiCo~\cite{maier2021heidelberg}, SurgYT \\
\bottomrule
\label{supp_1}
\end{tabular}
}
\end{table*}

\begin{table*}[ht]
\centering
\caption{Surgery-action distribution summary across six surgery types.}
\label{tab:action_surgery}
\small % 使用小字号以适应列宽
\setlength{\tabcolsep}{3.5pt} % 微调列间距
\resizebox{\linewidth}{!}{
\begin{tabular}{lccccccc}
\toprule
\textbf{Action} & \textbf{Cholecystectomy} & \textbf{Gastrectomy} & \textbf{Hysterectomy} & \textbf{Intestinal Resection} & \textbf{Nephrectomy} & \textbf{Prostatectomy} & \textbf{Total} \\
\midrule
Aspiration & 182 & 310 & 86 & 141 & 23 & 55 & 797 \\
Clipping & 668 & 27 & 12 & 10 & 103 & 132 & 952 \\
Coagulation & 205 & 280 & 26 & 69 & 23 & 32 & 635 \\
Dissection & 1562 & 2956 & 274 & 141 & 39 & 301 & 5273 \\
Knot-tying & 24 & 107 & 54 & 38 & 26 & 106 & 355 \\
Needle Grasping & 7 & 43 & 20 & 5 & 53 & 647 & 775 \\
Needle Puncture & 33 & 139 & 49 & 22 & 110 & 458 & 811 \\
Packaging & 90 & 11 & 7 & 5 & 25 & 11 & 149 \\
Suture Pulling & 32 & 59 & 31 & 8 & 54 & 623 & 807 \\
Tissue Retraction & 599 & 518 & 44 & 137 & 11 & 52 & 1361 \\
\midrule
\textbf{Total} & \textbf{3402} & \textbf{4450} & \textbf{603} & \textbf{576} & \textbf{467} & \textbf{2417} & \textbf{11915} \\
\bottomrule
\end{tabular}
}
\end{table*}

\begin{table*}[hb]
\centering
\setlength{\tabcolsep}{1mm}
\caption{Distribution of BSA classes across ten-fold cross-validation splits for the BSA-10 dataset.}
\label{tab:fold_act}

\begin{tabular}{cccccccccccc}
\hline
                 & fold1 & fold2 & fold3 & fold4 & fold5 & fold6 & fold7 & fold8 & fold9 & fold10 & TOTAL  \\ \hline
Aspiration       & 102   & 78    & 70    & 84    & 55    & 79    & 102   & 81    & 64    & 82     & 797       \\
Clipping         & 116   & 128   & 83    & 97    & 84    & 96    & 76    & 80    & 100   & 92     & 952       \\
Coagulation      & 81    & 98    & 38    & 126   & 30    & 90    & 41    & 41    & 57    & 33     & 635       \\
Dissection       & 389   & 689   & 473   & 599   & 498   & 625   & 650   & 502   & 446   & 402    & 5273      \\
Knot-tying         & 23    & 44    & 47    & 24    & 22    & 22    & 79    & 21    & 27    & 46     & 355       \\
Needle Grasping   & 58    & 66    & 73    & 102   & 97    & 82    & 50    & 85    & 88    & 74     & 775       \\
Needle Puncture   & 62    & 64    & 89    & 114   & 60    & 64    & 114   & 79    & 90    & 75     & 811       \\
Packaging          & 12    & 14    & 23    & 14    & 11    & 21    & 11    & 18    & 11    & 14     & 149       \\
Suture Pulling    & 78    & 92    & 86    & 110   & 82    & 64    & 76    & 57    & 78    & 84     & 807       \\
Tissue Retraction & 146   & 166   & 84    & 181   & 134   & 141   & 135   & 105   & 132   & 137    & 1361      \\ \hline
TOTAL             & 1067  & 1439  & 1066  & 1451  & 1073  & 1284  & 1334  & 1069  & 1093  & 1039   & 11915     \\ \hline
\end{tabular}

\end{table*}

\begin{table*}[ht]
\centering
\caption{Per-fold sample distribution across surgery types.}
\label{tab:fold_surg}
\resizebox{\linewidth}{!}{
\begin{tabular}{lrrrrrrrrrrr}
\toprule
Surgery Type & fold1 & fold2 & fold3 & fold4 & fold5 & fold6 & fold7 & fold8 & fold9 & fold10 & TOTAL  \\
\midrule
Cholecystectomy & 273 & 277 & 297 & 420 & 414 & 373 & 285 & 275 & 407 & 381 & 3402 \\
Gastrectomy     & 410 & 482 & 432 & 568 & 290 & 492 & 571 & 429 & 411 & 365 & 4450 \\
Hysterectomy    & 12 & 20 & 36 & 37 & 108 & 65 & 146 & 115 & 19 & 45 & 603 \\
Intestinal Resection & 66 & 88 & 25 & 29 & 19 & 137 & 174 & 18 & 5 & 15 & 576 \\
Nephrectomy     & 83 & 52 & 21 & 16 & 16 & 27 & 44 & 90 & 60 & 58 & 467 \\
Prostatectomy   & 223 & 520 & 255 & 381 & 226 & 190 & 114 & 142 & 191 & 175 & 2417 \\
\midrule
TOTAL           & 1067 & 1439 & 1066 & 1451 & 1073 & 1284 & 1334 & 1069 & 1093 & 1039 & 11915 \\
\bottomrule
\end{tabular}
}
\end{table*}

\begin{table*}[!htbp]
\centering
\caption{In lobectomy procedures, the model's performance across ten basic surgical action classes is quantified by the Youden Index (95\% CI).}

\begin{tabular}{ccccc}
\hline
Action            & AUROC                & Sensitivity          & Specificity          \\ \hline
Aspiration        & 90.62 (87.94, 93.31)   & 82.63 (75.75, 89.52)   & 90.45 (83.86, 97.04)   \\
Clipping          & 89.63 (85.32, 93.93)   & 85.45 (75.19, 95.71)   & 88.06 (82.74, 93.39)   \\
Coagulation       & 79.11 (75.14, 83.09)   & 80.00 (68.00, 92.00)   & 72.68 (62.01, 83.35)   \\
Dissection        & 88.29 (85.62, 90.97)   & 85.13 (81.47, 88.78)   & 79.09 (75.63, 82.55)   \\
Knot-tying          & 91.72 (87.12, 96.32)   & 85.00 (75.08, 94.92)   & 92.33 (87.73, 96.93)   \\
Needle Grasping   & 92.96 (89.48, 96.43)   & 88.33 (80.29, 96.38)   & 94.00 (90.38, 97.62)   \\
Needle Puncture   & 88.59 (83.23, 93.94)   & 78.00 (66.92, 89.08)   & 92.93 (88.83, 97.02)   \\
Packaging           & 93.77 (89.45, 98.09)   & 86.67 (77.26, 96.07)   & 93.96 (91.48, 96.44)   \\
Suture Pulling    & 87.01 (83.13, 90.90)   & 81.43 (71.73, 91.12)   & 86.36 (81.63, 91.08)   \\
Tissue Retraction & 80.63 (77.95, 83.30)   & 70.34 (61.94, 78.75)   & 80.66 (73.37, 87.95) 
  \\ \hline
\end{tabular}

\label{tab_supp_1}
\end{table*}

\begin{table*}[htbp]
\centering
\caption{In hepatectomy procedures, the model's performance across ten basic surgical action classes is quantified by the Youden Index (95\% CI).}
\label{tab:per_class_metrics}

\begin{tabular}{@{}lccc@{}}
\toprule
\textbf{Class} & \textbf{AUROC (95\% CI)} & \textbf{Sensitivity (95\% CI)} & \textbf{Specificity (95\% CI)} \\
\midrule
Aspiration & 88.30 (87.29, 89.32) & 74.32 (70.65, 78.00) & 91.04 (87.47, 94.61) \\
Clipping & 97.05 (96.24, 97.86) & 96.77 (93.90, 99.65) & 89.39 (86.35, 92.42) \\
Coagulation & 95.07 (93.35, 96.79) & 91.54 (88.41, 94.66) & 93.86 (90.22, 97.50) \\
Dissection & 94.68 (94.00, 95.36) & 87.56 (83.98, 91.14) & 91.64 (88.74, 94.54) \\
Knot-tying & 95.41 (92.21, 98.61) & 92.86 (85.63, 100.00) & 91.97 (87.72, 96.23) \\
Needle Grasping & 79.58 (73.90, 85.25) & 57.50 (42.78, 72.22) & 92.43 (85.98, 98.88) \\
Needle Puncture & 89.23 (85.13, 93.32) & 76.67 (67.15, 86.18) & 96.72 (95.69, 97.74) \\
Packaging & 97.07 (93.15, 100.00) & 95.00 (83.69, 100.00) & 97.21 (93.19, 100.00) \\
Suture Pulling & 91.90 (85.82, 97.98) & 85.00 (67.72, 100.00) & 93.32 (87.22, 99.41) \\
Tissue Retraction & 80.03 (78.15, 81.91) & 60.00 (51.86, 68.14) & 86.36 (79.77, 92.96) \\
\bottomrule
\end{tabular}

\label{tab_supp_2}
\end{table*}

%%%%%%%%%%% CAPTIONS FOR OTHER SUPPLEMENTARY FILES %%%%%%%%%%

\clearpage % Clear all remaining figures and tables then start a new page

\end{document}